\documentclass[11pt]{article}

\usepackage[preprint]{acl}

\usepackage{times}
\usepackage{latexsym}

\usepackage[T1]{fontenc}

\usepackage[utf8]{inputenc}

\usepackage{microtype}

\usepackage{inconsolata}

\usepackage{graphicx}

\usepackage{booktabs}
\usepackage{multirow}
\usepackage{amssymb}
\usepackage{arydshln}
\usepackage{amsmath}
\usepackage{algorithm}
\usepackage{algorithmic}

\title{CTTA-T: Continual Test-Time Adaptation for Text Understanding \\ via Teacher-Student with a Domain-aware and Generalized Teacher}

\author{
  \textbf{Tianlun Liu\textsuperscript{1}},
  \textbf{Zhiliang Tian\textsuperscript{1}},
  \textbf{Zhen Huang\textsuperscript{1}},
  \textbf{Xingzhi Zhou\textsuperscript{2}}, \\
  \textbf{Wanlong Yu\textsuperscript{1}},
  \textbf{Tianle Liu\textsuperscript{1}},
  \textbf{Feng Liu\textsuperscript{1}},
  \textbf{Dongsheng Li\textsuperscript{1}}
\\
\\
  \textsuperscript{1}College of Computer Science and Technology, National University of Defense Technology, Hunan, China
\\
  \textsuperscript{2}The Hong Kong University of Science and Technology, Hong Kong, China
\\
  \small{
    \{ltlun, tianzhiliang, huangzhen, ywl, liutianle, richardlf, dsli\}@nudt.edu.cn \small\{xzhoubl\}@cse.ust.hk
  }
}

\begin{document}
\maketitle
\begin{abstract}
Text understanding often suffers from domain shifts.
To handle testing domains, domain adaptation (DA) is trained to adapt to a fixed and observed testing domain; a more challenging paradigm, test-time adaptation (TTA), cannot access the testing domain during training and online adapts to the testing samples during testing, where the samples are from a fixed domain. We aim to explore a more practical and underexplored scenario, continual test-time adaptation (CTTA) for text understanding, which involves a sequence of testing (unobserved) domains in testing.
Current CTTA methods struggle in reducing error accumulation over domains and enhancing generalization to handle unobserved domains: 1)~Noise-filtering reduces accumulated errors but discards useful information, and 2) accumulating historical domains enhances generalization, but it is hard to achieve adaptive accumulation. 
In this paper, we propose a \textbf{CTTA-T} (\textbf{c}ontinual \textbf{t}est-\textbf{t}ime \textbf{a}daptation for \textbf{t}ext understanding) framework adaptable to evolving target domains: it adopts a teacher-student framework, where the teacher is domain-aware and generalized for evolving domains.
To improve teacher predictions, we propose a refine-then-filter based on dropout-driven consistency, which calibrates predictions and removes unreliable guidance.
For the adaptation–generalization trade-off, we construct a domain-aware teacher by dynamically accumulating cross-domain semantics via incremental PCA, which continuously tracks domain shifts. Experiments show CTTA-T excels baselines. 

\end{abstract}

\section{Introduction}
\label{sec:1}

Text understanding has a wide range of applications. Most existing models rely on offline training with fixed source data and perform well when the testing and training domains are aligned \cite{devlin-etal-2019-bert,ijcai2018p0570}. However, in real-world text understanding applications, testing domains shift over time, a phenomenon known as continual domain shift \cite{wang2022continual}. 
This occurs, for instance, when a model trained in a source domain (i.e., movie reviewing) is later deployed for inference on various target domains (i.e., restaurant reviewing or book reviewing). Most current models do not improve themselves as they are applied to new target domains. It is sometimes necessary for models to continually enhance themselves as they encounter evolving target domains.

\begin{figure}[t]
    \centering
    \includegraphics[width=1\columnwidth]{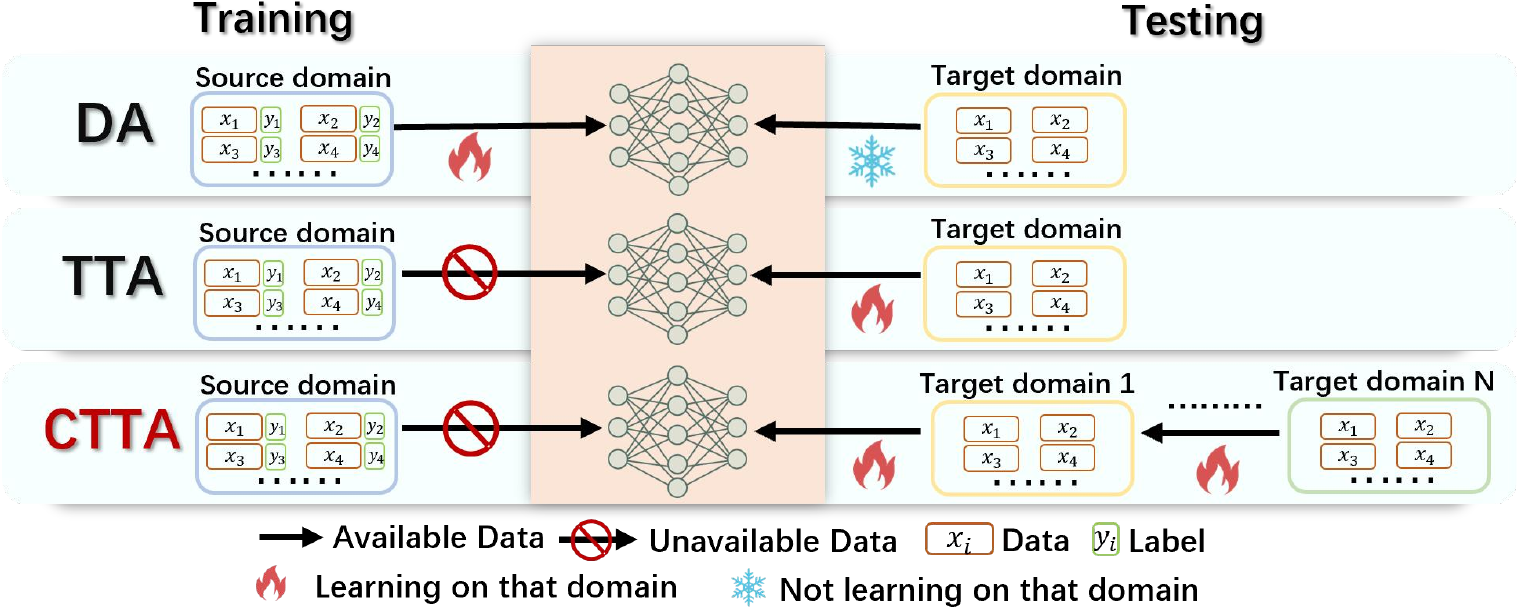}  
    \caption{
        Comparison of three settings for adapting to domain shift. DA, TTA, and CTTA represent progressively more challenging and practical scenarios. 
    }
    \label{fig:1234}
\end{figure}
To mitigate the domain shift, some researchers explore domain adaptation (DA) in text understanding. DA improves model performance by reducing the gap between source (training) and target (testing) domains \cite{cao2020unsupervised} to mitigate a single domain shift, which requires observed information of the target domain during training.
However, in practice, domain shifts are often observed during testing, which limits the applicability of DA.
A new task, test-time adaptation (TTA), enables models to adapt to domain shift during testing by learning online on unlabeled testing data from the fixed target domain.
However, real-world scenarios always involve continual domain shifts (i.e., continuous multiple shifts), where domains shift over time, and TTA methods are inadequate in handling such scenarios. To overcome the limitation, we study continual test-time adaptation (CTTA) in text understanding, where unlabeled testing samples arrive sequentially from continuously shifting domains.
In this setting, the model must predict and adapt online without access to training samples or past and future testing samples. The illustration of the three settings (DA, TTA, and CTTA) is in Fig.~\ref{fig:1234}.


Although TTA has explored techniques such as entropy minimization \cite{wang2021tent}, teacher-student self-training \cite{lee2013pseudo, ye2022robust}, and contrastive learning \cite{chen2022contrastive}, these methods are not fully applicable under CTTA for text understanding due to two limitations:
1) continual domain shift amplifies noise accumulation over evolving domains, eventually causing collapse \cite{guo2017calibration, chen2019progressive}. 
2)~TTA focuses on fixed domain shifts but lacks mechanisms to adapt to evolving textual patterns for evolving domains, leading to poor generalization.

To fix noise accumulation and generalization issues in CTTA mentioned above, researchers propose sample filtering and teacher-student framework methods:
(1) To mitigate noise accumulation, researchers proposed sample filtering methods \cite{leeentropy, wang-2024-continual} that remove noisy samples during adaptation to improve robustness. However, these methods often struggle to distinguish noise from useful signals, and directly filtering samples may harm performance by discarding target-domain information.
(2) To address generalization issues, researchers proposed a teacher-student framework \cite{karim2023c,lyu2024variational} that accumulates information from previous domains to build a teacher for generating pseudo-labels to guide the student. This enhances generalization across domains: the teacher, updated by partially absorbing the student's parameters, provides cross-domain understanding ability, while the student adapts to each specific domain.
When teacher accumulates historical domain information from student, the contribution of each domain remains fixed at every update and cannot adapt dynamically to domain-specific semantic information.

In summary, when applying existing CTTA methods (i.e., teacher-student with sample filtering) to text understanding, models struggle to balance the useful and noisy information from the target domain. We argue that we should carefully and adaptively absorb the sample-level and domain-level information: (1) considering uncertainty in sample filtering, and (2) dynamically accumulating domain-specific semantic information.

In this paper, we propose the \textbf{c}ontinual \textbf{t}est-\textbf{t}ime \textbf{a}daptation for
\textbf{t}ext understanding \textbf{(CTTA-T)} framework: 
It adopts a teacher-student framework, equipping the teacher with both domain awareness and generalization to adapt to evolving domains, while applying a consistency-based refine-then-filter module to improve the quality of its predictions.
Specifically, we use the teacher-student framework as the foundation for our implementation in CTTA for text understanding. 
We propose a refine-then-filter module to improve teacher prediction by leveraging consistency from multiple dropout forward passes. It first refines the teacher’s outputs via consistency-based reweighting, then filters low-consistency predictions.
We propose a domain-aware teacher update module that employs incremental principal component analysis (IPCA) to compute semantic shift, enabling the teacher to perform cross-domain accumulation by dynamically adjusting its integration of target-domain information, balancing specific-domain adaptation and cross-domain generalization.
We stochastically restore part of the teacher’s parameters, injecting source general knowledge to enhance its generalization.
Since there is currently no available benchmark for CTTA in text understanding, we further construct a benchmark that covers multiple domains and task sequences across robust QA, reading comprehension, cross-lingual QA, and sentiment analysis.
Experiments show that, unlike all baselines whose performance degrades under CTTA, our method maintains more stable performance and outperforms all baselines.

Our contributions are: \textbf{(1)} We propose \textbf{CTTA-T}, a continual test-time adaptation (CTTA) framework for text understanding, and introduce a benchmark tailored for evaluating CTTA.  
\textbf{(2)} We introduce a refine-then-filter module to improve teacher prediction reliability based on prediction consistency. 
\textbf{(3)} We propose a domain-aware teacher update module that applies dynamically cross-domain accumulation via IPCA to balance adaptation and generalization.
\textbf{(4)} Experiments on the CTTA benchmark demonstrate that CTTA-T achieves SOTA.


\section{Related Work}
\textbf{Domain Adaptation}  
aims to improve generalization across domains in text understanding. It falls into two types:
\textbf{1) Data-based methods} adapt to the target domain by generating or selecting relevant data.
CANMD \cite{yue2022contrastive} generates high-confidence pseudo-labels.
T-SAS \cite{jeong-etal-2023-test} mitigates noise by generating and filtering answers.
\textbf{2) Self-training-based methods} mitigate domain gaps by modifying model architecture or training protocols.
TTT-NN \cite{hardt2024testtime} retrieves neighbors during testing.
CrossIn \cite{lin-etal-2025-crossin} enables cross-lingual tuning using translated pairs.
However, DA assumes access to domain shifts during training, while in practice, such shifts are observable at testing.

\textbf{Test-Time Adaptation}  
is a testing process that mitigates domain shift \cite{wang2021tent}. It falls into two types:
\textbf{1) Non-backpropagation-based methods} update modules or outputs solely during testing. 
LAME \cite{boudiaf2022parameter} adjusts probabilities with Laplace maximum likelihood.
FOA \cite{niu2024test} uses covariance adaptation to learn input prompts.
\textbf{2) Backpropagation-based methods} update the model by optimizing using test samples. 
Tent \cite{wang2021tent} adapts during testing by minimizing entropy.
CoTTA \cite{wang2022continual} uses a teacher-student framework. 
OIL \cite{ye2022robust} uses a teacher-student framework and causal inference.
SAR \cite{niu2023towards} conducts sharpness-aware entropy
minimization.
SoTTA \cite{gong2023sotta} buffers class-balanced samples and sharpens entropy minimization.
REM \cite{han2025ranked} builds the prediction difficulty via a progressive masking strategy.
Existing work focuses primarily on computer vision (CV), and our goal is to achieve strong performance on continual domain shift in text understanding.

\section{Method}\label{sec:3}
\begin{figure*}[t]
    \centering
    \includegraphics[width=0.97\textwidth]{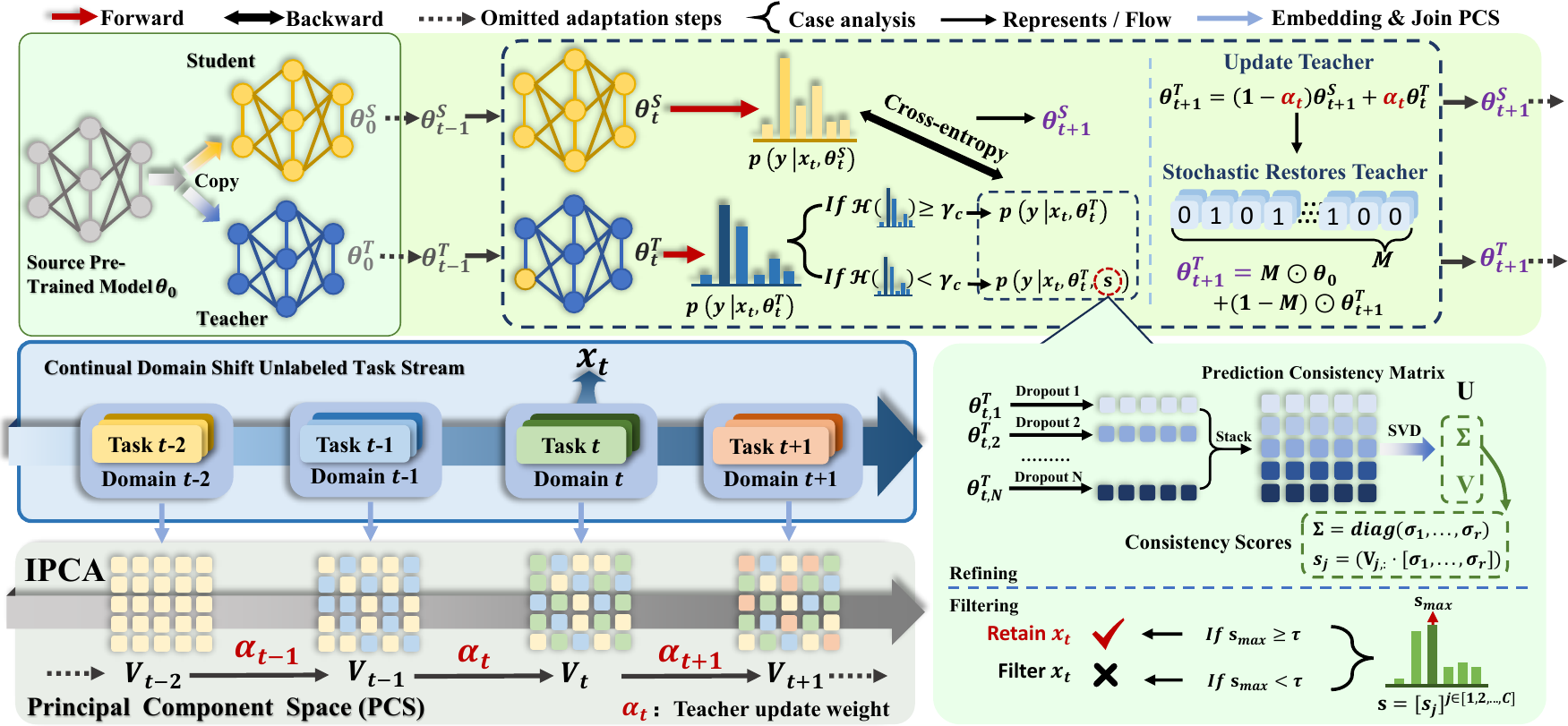}  
    \caption{
        Overview of \textbf{CTTA-T}. 
When a textual sample \(x_t\) arrives (blue box), teacher and student predict on it. Low-entropy teacher outputs guide the student via cross-entropy; otherwise, \(\mathbf{s}_{\text{max}}\) is computed (bottom-right). Low-\(\mathbf{s}_{\text{max}}\) samples are discarded, while high ones combine the consistency distribution with the teacher output for refined student update. The student then updates the teacher (upper top-right), with update weight measured by IPCA domain distance (bottom-left). Finally, part of the teacher’s parameters is randomly restored (lower top-right).
    }
    \label{fig:main}
\end{figure*}

The CTTA-T framework comprises four modules as shown in Fig.~\ref{fig:main}. 
(1) We employ a teacher-student framework as the foundation of our CTTA approach for text understanding (\S \ref{sec:3.1}): the teacher absorbs cross-domain information, and students learn domain-specific information. 
(2) We apply a refine-then-filter module (\S \ref{sec:3.2}) via consistency to improve teacher prediction reliability, which reduces the teacher-predicted uncertainty according to its consistency.
(3) We propose a domain-aware teacher module (\S \ref{sec:3.3}) that dynamically accumulates cross-domain semantic information from the student. We adopt incremental principal component analysis (IPCA) to extract and update the semantic space that reflects the evolving domain information.
(4) We introduce a stochastic restoration module (\S \ref{sec:3.4}) to enhance teacher generalization by randomly injecting source-domain knowledge.

\noindent\textcolor{black}{
\textbf{Problem Definition:} Given a model pre-trained on source domain $P_s$, we aim to adapt it to a continually shifting target domain $P_t$. The model updates using the no-label of the current test data $x_t^T$, without accessing source data \(X^s\), past test data \(X_{<t}^T\), or future test data \(X_{>t}^T\). (Details are in App.~\ref{app:Problem Definition}.)}  

\subsection{Teacher-Student Model for CTTA in Text Understanding}
\label{sec:3.1}
To achieve continual test-time adaptation (CTTA) for text understanding, we build a CTTA backbone using a teacher-student framework where the teacher continuously guides the student during testing. This is motivated by text understanding, which involves context-sensitive semantics and domain-specific terminology,
making continual adaptation crucial as domains shift during testing.

Let $\theta_{t}^{T}$ and $\theta_{t}^{S}$ denote the teacher and student models at time $t$, both initialized from the source model $\theta_0$. At each time, we enforce consistency between their predictions by minimizing the cross-entropy of their softmax probability outputs:
\begin{equation}\small
\label{eq:1}
\mathcal{L}_{\theta_t^S}(x_t)=-\sum_y p\left(y|x_t,\theta_t^T\right)\log p\left(y|x_t,\theta_t^S\right).
\end{equation}

After the student is updated via Eq.~\ref{eq:1}, the teacher updates by partially absorbing the student with a fixed weight.
In this framework, the student model $\theta_{t}^{S}$ acts as a fast adapter to the current domain, capturing domain-specific semantics from the textual embedding of $x_t$ by directly updating its parameters via Eq.~\ref{eq:1}. The teacher model $\theta_{t}^{T}$, in contrast, serves as a stable knowledge aggregator, progressively integrating the student’s updates to maintain a general understanding of semantics across testing domains. This allows the teacher to provide reliable pseudo-labels to guide the student, even as the input domain shifts continuously.


\subsection{Refine-then-Filter Teacher Prediction via Consistency}
\label{sec:3.2}

We propose a \textit{Refine-then-Filter Teacher Prediction} (RFP) module that enhances teacher guidance by improving the reliability of teacher predictions before using them to guide the student. 
In CTTA, adaptation under continual domain shifts amplifies two types of uncertainty: \textit{epistemic uncertainty} (EU), caused by model uncertainty, which leads to unstable predictions on unseen domains, \textit{aleatoric uncertainty} (AU), caused by input noise, which introduces random noise even for familiar domain inputs \cite{kendall2017uncertainties}.
As the EU stems from insufficient model confidence and the AU from irreducible data noise, handling them jointly is ineffective.
To mitigate their effects, our RFP operates in two stages: the \textbf{refine} stage reduces EU by leveraging prediction consistency (i.e., the agreement among multiple stochastic forward passes under our dropout) to calibrate the teacher’s outputs, which stabilizes predictions in unfamiliar domains.
The \textbf{filter} stage suppresses AU by removing predictions with low consistency, as these likely arise from intrinsically noisy; eliminating them prevents such randomness from passing to the student.
Hence, explicitly modeling EU and AU becomes essential to prevent accumulated uncertainty from destabilizing continual adaptation in CTTA.

\noindent{\textbf{Refinement via Prediction Consistency Distribution.}}
To identify samples with high EU, we first use entropy as a screening mechanism and then perform refinement based on prediction consistency. We compute the entropy of the teacher’s prediction as a proxy to detect high-EU samples, following established practice \cite{kendall2017uncertainties}. 
Specifically, for a given sample $x_t$, we compute the entropy $\mathcal{H}(p(y|x_t,\theta_t^T))$ using $\theta_t^T$, and mark it as high-EU if the entropy exceeds a threshold $\gamma_c$.
Once these high-EU samples are identified, we introduce a three-step prediction refinement process:

\textbf{Step 1: Construct the prediction consistency matrix.}
We take $N$ stochastic dropout masks on the teacher model $\theta^T_t$ at time $t$, yielding $N$ sampled models ${\theta_n^{dp}}^{n\in[1,2,...,N]}$. Each $\theta_n^{dp}$ belongs to a variational distribution $q(\theta)$ that approximates the true posterior. This enables MC sampling to approximate the model’s predictive distribution and quantify EU through the variability of predictions across different dropout-induced parameter samples.
For input $x_t$, we obtain the probability $p_n = P(y|x_t, \theta_n^{dp})$ from each sampled model. We then construct a prediction consistency matrix $P_C$ by stacking these probability vectors column-wise:
$P_C=\left[p_1,...,p_N\right]^T\in\mathbb{R}^{N\times C}$, 
where each row corresponds to one stochastic forward pass and each column corresponds to a class. Finally, we apply SVD to $P_C$ to capture the principal consistency patterns among predictions:
\begin{equation}
    P_C=U\Sigma V^T,\quad\Sigma=\mathrm{diag}(\sigma_1,\sigma_2,...,\sigma_r),
\end{equation}
where \( r = \text{rank}(P_C) \), 
Each row of $V^T$ represents a principal direction in the class space, whose corresponding singular value \( \sigma_i \) quantifies its importance.
A large \( v_{ji} \) indicates a strong correlation between class \( j \) and principal direction \( i \). 
By constructing and decomposing $P_C$, the model identifies coherent uncertainty patterns across MC samples, thereby isolating EU from unstable model beliefs.

\textbf{Step 2: Obtain prediction consistency distribution.}
We define the prediction consistency distribution $\mathbf{s}$ as a class-wise summary of these principal components \( V_{:,i} \). For each class $j$, its score $\mathbf{s}_j$ is computed as the weighted sum of the principal directions $\{V_{:,i}\}_{i=1}^{r}$ associated with that class, using singular values $\{\sigma_i\}_{i=1}^{r}$ as weights:
\begin{equation}
\label{eq:prediction_consistency_score}
    \mathbf{s} = [\mathbf{s}_j]^{j\in[1,2,\ldots,C]},
\end{equation}
where $\mathbf{s}_j = \sum_{i=1}^{r} V_{j,i}\sigma_i$, and $V_{j,:}$ denotes the $j$-th row of $V$.
The singular values capture how dominant each principal direction is among all stochastic predictions, 
while $V_{j,:}$ measures how much class $j$ aligns with these stable directions. 
Therefore, a higher $\mathbf{s}_j$ indicates that class $j$ consistently receives similar predictions across multiple 
MC-dropout samples, corresponding to lower EU.



\textbf{Step 3: Refine the sample with prediction consistency distribution.}
We define the refined probability $p(y|x_t,\theta^T_t,\mathbf{s})$ as Eq.~\ref{eq:refined_softmax_probabilities}, 
where the prediction consistency distribution $\mathbf{s}$ is softmax-transformed to re-weight the original probabilities.
\begin{equation}\small
 \label{eq:refined_softmax_probabilities}
    p(y|x_t,\theta^T_t,\mathbf{s}) = \text{softmax}({\text{softmax}(\mathbf{s})\odot p(y|x_t,\theta_t^T))},
\end{equation} 
where $\odot$ is Hadamard product. For high-EU samples, we replace teacher’s prediction in Eq. \ref{eq:1} with refined probability $p(y|x_t,\theta^T_t,\mathbf{s})$ to reduce EU.

\noindent{\textbf{Filtering via Prediction Consistency Score.}}
Since AU originates from data noise and cannot be directly mitigated, we apply a filter guided by the prediction consistency score $\mathbf{s}$ to reduce AU. 
Specifically, we filter out samples if its prediction consistency score $\mathbf{s}_{\text{max}}=\max_j [\mathbf{s}_j]^{j\in[1,2,\ldots,\mathcal{C}]}$ is below the threshold $\tau$, 
where $[\mathbf{s}_j]^{j\in[1,2,\ldots,\mathcal{C}]}$ (as eq. \ref{eq:prediction_consistency_score}) is the prediction consistency distribution.
When the sample $x_t$ is dominated by data noise, the consistency matrix \( P_C \) exhibits random behavior and causes a drop in \( \mathbf{s}_{\text{max}}\). 
Therefore, setting a threshold $\tau$ to filter out smaller $\mathbf{s}_{\text{max}}$ enables effective masking of samples with high AU (The proof of the upper bound of consistency score is in App.~\ref{app:c.1}).
\subsection{Cross-Domain Accumulation for Domain-Aware Teacher}

\label{sec:3.3}
We propose a \textit{Cross-Domain Accumulation for Domain-Aware Teacher (CDA)} module that enables a domain-aware teacher to dynamically accumulate cross-domain semantic information for text understanding in CTTA. 
Most teacher-student CTTA models adopt a fixed-weight exponential moving average (EMA) update \cite{tarvainen2017mean}. We argue that a domain-aware teacher should dynamically absorb student knowledge based on cross-domain shifts. 
Traditional EMA updates the teacher $\theta_t^T$ at each time $t$ as:
\begin{equation}
\label{eq:2}
\theta_{t+1}^T=\alpha\theta_t^T+(1-\alpha)\theta_{t+1}^S,
\end{equation}
where the $\theta_{t+1}^{S}$ is updated student model through Eq. \ref{eq:1}. 
However, traditional EMA, using a static accumulation rate $\alpha$, fails to reflect domain shifts in streaming data, leading to either over-adaptation or under-adaptation in evolving text distributions.
To facilitate domain-aware accumulation, CDA dynamically adjusts the update rate $\alpha$ by measuring the semantic distance between all historical and current domains via incremental principal component analysis (IPCA) \cite{2008Incremental} in two steps: 

\textbf{Step 1: Measuring domain semantic distance.}
We use IPCA to extract low-dimensional semantic representations \( C \) and measure the semantic shift distance between the current and historical domains.
The motivation for using IPCA lies in its ability to incrementally accumulate domain semantics into a compact representation, enabling continual tracking of cross-domain evolution without storing past data.
Specifically, we treat the covariance matrix as a proxy for domain-level semantics, since it captures co-occurrence patterns among features, which reflect the structural semantics of domain.
At each time \( t \), we compute the covariance matrix \( C_t \) from all historical samples \( X_{\leq t} \), and apply singular value decomposition (SVD) to obtain the principal component space \( V_t \) (\( C_t = V_t \Lambda_t V_t^\top \)). When new data \( X_{t+1} \) arrives, IPCA integrates the new domain’s semantic statistics into the historical covariance matrix, updating \( C_{t} \) to \( C_{t+1} \) as follows (see App.~\ref{app:b.1} for detailed IPCA derivation):
\begin{equation}\label{eq:3}\small
C_{t+1}\approx\frac{N_tC_t+n\hat{C}+\frac{N_tn}{N_t+n}\left(\mu_t-\hat{\mu}\right)\left(\mu_t-\hat{\mu}\right)^\top}{N_t+n},
\end{equation}
where \( N_t \), \( \mu_t \) denote the count and mean of historical samples, and \( n \), \( \hat{\mu} \), \( \hat{C} \) are the numbers, mean, and covariance of current samples.
This allows us to incrementally maintain an up-to-date semantic representation \( C_{t+1} \) without storing all samples.
The corresponding principal component matrices \( V_t \) and \( V_{t+1} \) represent the main semantic axes of the historical and current domains. Since the principal component space captures the densest semantic directions, the Frobenius norm of their variation,
\begin{equation}\label{eq:4}
distance \approx \lVert \Delta V \rVert_F^2, 
\end{equation}
where \( \Delta V = V_t - V_{t+1} \), measures how much the domain semantics have shifted over time (see App. \ref{app:b.2} for $distance$ detailed derivation). The Frobenius norm summarizes the squared difference of each component axis, offering an intuitive and efficient measure of domain-level variation.
In conclusion, IPCA allows us to accumulate cross-domain semantics in a low-dimensional space and dynamically quantify domain shift over time, forming the basis for domain-aware teacher adaptation in CDA.

\textbf{Step 2: Obtain teacher weight $\alpha$ according to domain semantic distance.}
We obtain the teacher’s update weight $\alpha$ based on the semantic distance measured in Step 1.
Intuitively, a large shift distance (large $\Delta V$) indicates that the current domain introduces substantial semantic change. To address it, the teacher needs to obtain more up-to-date domain information provided by the student ($\theta_{t+1}^S$) to adapt effectively.
Conversely, small $\Delta V$ implies high semantic similarity between current and historical domains, so retaining prior knowledge and relying more on previous parameters ($\theta_{t}^T$) is better.
To formalize the relation, we define $\mathcal{F}$ that reflects the trade-off between aligning with the current student and preserving past information:
\begin{equation}\label{eq:5}
\begin{aligned}
\mathcal{F}(\theta_{t+1}^T)=\parallel\Delta V\parallel_F^2&\cdot(\theta_{t+1}^T-\theta_{t+1}^S)^2\\
+(1-\parallel\Delta V\parallel_F^2)&\cdot(\theta_{t+1}^T-\theta_t^T)^2.
\end{aligned}
\end{equation}
It shows that when the distance between all historical domains and the current domain increases (i.e., large $\Delta V$), the term $(\theta_{t+1}^T - \theta_{t+1}^S)^2$ is emphasized, forcing the teacher to align with the student and adapt to the new domain. When $\Delta V$ is small, 
the term $(\theta_{t+1}^T - \theta_t^T)^2$ is emphasized, 
forcing the teacher to retain historical domains information by pushing teacher to its previous parameters.
\textcolor{black}{Then, we differentiate Eq.~\ref{eq:5} to obtain the gradient for updating the teacher model. Finally, we combine the result of the differentiation with Eq. \ref{eq:2} to derive the weight as \( \alpha = 1 - \textit{distance}\). To stabilize the model, we map $\alpha$ to a bounded range. (derivation in App.~\ref{app:b.4}). 
The CDA provides a theoretical mechanism to replace the fixed $\alpha$ with a domain-aware update rule.}

When \( \Delta V \) is large, the update weight \( \alpha \) decreases, prompting the teacher to absorb more from the student and better adapt to the current domain; vice versa, preserving historical parameters to prevent overfitting and maintain general information.
CDA continuously adjusts $\alpha$ to modulate the trade-off between general knowledge retention and new domain-specific knowledge in CTTA.
\subsection{Stochastic Restoration for Generalized Teacher}
\label{sec:3.4}
We propose a \textit{Stochastic Restoration for Generalized Teachers (SRT)} module that enhances the generalization ability of the teacher model by randomly restoring a portion of its parameters to their initial source values.
The motivation is to enhance the generalization of the teacher by injecting source-domain knowledge, thereby preventing it from overfitting to transient and potentially noisy test-time domains.
Specifically, for each parameter tensor, we independently sample a binary mask $M \in \{0,1\}^d$ from a Bernoulli distribution with restoration probability $p$, i.e., $M \sim \text{Bernoulli}(p)$. 
The restoration operation is defined as:
\begin{equation}
\theta_{t+1}^T = M \odot \theta_0 + (1 - M) \odot \theta_{t+1}^T,
\end{equation}
where $\odot$ denotes Hadamard product, $\theta_0$ is source parameters, and $\theta_{t+1}^T$ is teacher parameter updated via Eq.~\ref{eq:2}.
By partially restoring the teacher's parameters, 
SRT constrains the deviation from general representations to improve generalization.

Note that SRT is compatible with our cross-domain accumulation module (CDA), while CDA adapts the teacher to current domain, SRT injects source generalization signals, jointly balancing adaptation and generalization.
Inspired by CoTTA \cite{wang2022continual}, SRT mitigates overfitting, but differs in a key design choice from CoTTA: restoration is applied to teacher rather than student. This avoids disrupting inference-critical parameters and enhances the teacher’s generalization.
\begin{table*}[h]
    \centering
    \resizebox{0.98
    \textwidth}{!}{
    \begin{tabular}{lcccccccccccccc}
        \toprule
        \multicolumn{1}{l}{} & \multicolumn{6}{c}{\textbf{Short Sequence}} & \multicolumn{6}{c}{\textbf{Long Sequence}} &  \\
        \midrule
         \textbf{Order}& \multicolumn{2}{c}{\textbf{Order1 (\textbf{t}$\longrightarrow$)}} & \multicolumn{2}{c}{\textbf{Order2 (\textbf{t}$\longrightarrow$)}} & \multicolumn{2}{c}{\textbf{Order3 (\textbf{t}$\longrightarrow$)}} & \multicolumn{2}{c}{\textbf{Order4} (\textbf{t}$\longrightarrow$)} & \multicolumn{2}{c}{\textbf{Order5} (\textbf{t}$\longrightarrow$)} & \multicolumn{2}{c}{\textbf{Order6} (\textbf{t}$\longrightarrow$)} &\multicolumn{2}{c}{\textbf{Avg}}  \\
        \midrule
         Methods & EM & F1 & EM & F1 & EM & F1 & EM & F1 & EM & F1 & EM & F1 & EM & F1 \\
        \midrule
        base & 52.52 & 65.45 & 52.52 & 65.45 & 52.52 & 65.45 & 55.94 & 68.27 & 55.94 & 68.27 & 55.94 & 68.27 & 54.23 & 66.86 \\
        \hdashline\noalign{\vskip 0.5mm}
        Tent & 28.83 & 34.43 & 29.66 & 39.89 & 38.34 & 51.23 & 36.04 & 43.63 & 19.69 & 27.19 & 8.27 & 12.20 & 26.81 & 34.76 \\

        OIL & 39.53 & 53.31 & 36.89 & 47.69 & 46.60 & 56.55 & 37.77 & 44.43 & 43.48 & 53.42 & 28.07 & 35.09 & 38.72 & 48.42 \\
        CoTTA & \underline{53.81} & \underline{65.75} & 44.19 & 55.74 & 49.15 & 61.01 & 47.65 & 57.68 & 23.77 & 31.84 & 15.99 & 22.14 & 39.09 & 49.03 \\
            SAR & 51.80 & 61.68 & \underline{52.13} & \underline{64.50} & \underline{52.52} & \underline{64.69} & \underline{53.08} & \underline{65.10} & \underline{52.77} & \underline{64.55} & \underline{50.36} & \underline{62.42} & \underline{52.11} & \underline{63.82} \\
                SoTTA & 35.23 & 45.67 & 45.43 & 56.60 & 48.83 & 60.20 & 47.37 & 58.56 & 38.15 & 50.20 & 30.25 & 41.97 & 40.88 & 52.20 \\
                        REM & 39.45 & 49.16 & 44.31 & 55.25 & 44.93 & 54.07 & 45.08 & 54.63 & 35.28 & 44.94 & 34.55 & 46.31 & 40.60 & 50.73 \\
        \textbf{CTTA-T (ours)} & \textbf{54.14} & \textbf{66.57} & \textbf{53.88} & \textbf{66.01} & \textbf{54.07} & \textbf{66.12} & \textbf{57.74} & \textbf{69.33} & \textbf{57.63} & \textbf{69.33} & \textbf{57.71} & \textbf{69.24} & \textbf{55.86} & \textbf{67.77} \\
        \midrule
        xTune & 56.96 & 68.98 & 56.96 & 68.98 & 56.96 & 68.98 & 59.93 & 71.44 & 59.93 & 71.44 & 59.93 & 71.44 & 58.45 & 70.21 \\
        \hdashline\noalign{\vskip 0.5mm}
        Tent & 50.72 & 62.13 & 44.79 & 56.64 & 44.52 & 54.47 & 39.06 & 49.06 & 23.06 & 29.26 & 55.91 & 67.63 & 43.01 & 53.20 \\

        OIL & 44.59 & 57.01 & 48.78 & 60.34 & 54.37 & 65.64 & 41.08 & 48.04 & 51.00 & 61.19 & 39.20 & 47.28 & 46.50 & 56.58 \\
        CoTTA & 51.49 & 63.57 & 53.24 & 65.77 & 52.02 & 64.18 & 54.80 & 66.70 & 55.57 & \underline{67.61} & \underline{56.28} & \underline{68.04} & 53.90 & 65.98 \\
            SAR & \underline{55.50} & \underline{66.86} & 56.19 & \underline{67.54} & \underline{56.15} & \underline{67.56} & \underline{58.10} & \underline{69.02} & \underline{55.59} & 66.78 & 49.99 & 61.21 & \underline{55.25} & \underline{66.50} \\
                SoTTA & 42.41 & 53.79 & \underline{56.34} & \underline{67.54} & 54.76 & 65.89 & 54.72 & 65.71 & 47.04 & 57.79 & 45.91 & 56.62 & 50.20 & 61.22 \\
                 REM & 49.72 & 61.07 & 49.72 & 62.24 & 51.57 & 62.82 & 52.34 & 63.76 & 50.69 & 62.15 & 45.94 & 58.77 & 50.00 & 61.80  \\  
        \textbf{CTTA-T (ours)} & \textbf{57.85} & \textbf{69.72} & \textbf{57.99} & \textbf{69.76} & \textbf{58.00} & \textbf{69.80} & \textbf{60.84} & \textbf{72.08} & \textbf{60.92} & \textbf{72.18} & \textbf{60.86} & \textbf{71.96} & \textbf{59.41} & \textbf{70.92} \\
        \bottomrule
    \end{tabular}
    }
    \caption{
    Main results (\%). We report EM and F1, and the final column shows the average score across all orders. \textbf{Bold}: the best. \underline{Underline}: the second-best. “t $\rightarrow$”: the adaptation on a temporally ordered task stream. 
    All results are statistically significant based on the t-test ($p<0.01$), details in App.~\ref{app:t-test}. Hyperparameter analysis is in App.~\ref{app:Hyperpara}
    }
    \label{tab: main_table}
\end{table*}
\begin{table*}[h]
  \centering
    \resizebox{1\textwidth}{!}{  
    \begin{tabular}{lcccccccccccccccccc}
    \toprule
    \multicolumn{5}{l}{} & \multicolumn{6}{c}{\textbf{Short Sequence}} & \multicolumn{6}{c}{\textbf{Long Sequence}} &  \\
    \midrule
     \multicolumn{1}{l}{Order} & TS-CTTA & CDA & RFP & SRT & \multicolumn{2}{c}{\textbf{Order1 (t$\longrightarrow$)}} & 
     \multicolumn{2}{c}{\textbf{Order2 (t$\longrightarrow$)}} & 
     \multicolumn{2}{c}{\textbf{Order3 (t$\longrightarrow$)}} & 
     \multicolumn{2}{c}{\textbf{Order4 (t$\longrightarrow$)}} & 
     \multicolumn{2}{c}{\textbf{Order5 (t$\longrightarrow$)}} & 
     \multicolumn{2}{c}{\textbf{Order6 (t$\longrightarrow$)}} &
     \multicolumn{2}{c}{\textbf{Avg}} \\
    \midrule   
    \multicolumn{1}{l}{base} & & & & & EM & F1 & EM & F1 & EM & F1 & EM & F1 & EM & F1 & EM & F1 & EM & F1 \\
    \midrule
    A & \textbf{\checkmark} &       &       &       & 37.24  & 48.11  & 44.72  & 56.44  & 49.32  & 60.93  & 40.83  & 51.05  & 42.18  & 52.18  & 29.81  & 37.83  & 40.68  & 51.09  \\
    B & \textbf{\checkmark} & \textbf{\checkmark} &       &       & 45.91  & 56.17  & 47.95  & 59.05  & 49.79  & 60.81  & 44.48  & 53.56  & 47.89  & 58.75  & 35.67  & 45.15  & 45.28  & 55.58  \\
    C & \textbf{\checkmark} & \textbf{\checkmark} & \textbf{\checkmark} &       & 48.12  & 59.35  & 46.37  & 57.29  & 51.22  & 62.78  & 48.71  & 58.74  & 51.90  & 63.17  & 39.46  & 49.76  & 47.63  & 58.52  \\
    \textbf{CTTA-T (ours)} & \textbf{\checkmark} & \textbf{\checkmark} & \textbf{\checkmark} & \textbf{\checkmark} & \textbf{54.14} & \textbf{66.57} & \textbf{53.88} & \textbf{66.01} & \textbf{54.07} & \textbf{66.12} & \textbf{57.74} & \textbf{69.33} & \textbf{57.63} & \textbf{69.33} & \textbf{57.71} & \textbf{69.24} & \textbf{55.86} & \textbf{67.77} \\
    \midrule
    xTune & & & & & EM & F1 & EM & F1 & EM & F1 & EM & F1 & EM & F1 & EM & F1 & EM & F1 \\
    \midrule
    A & \textbf{\checkmark} &       &       &       & 46.43 & 55.97  & 47.74  & 58.03  & 53.33  & 63.86  & 45.02  & 53.62  & 46.15  & 54.35  & 37.38  & 47.41  & 46.01  & 55.54  \\
    B & \textbf{\checkmark} & \textbf{\checkmark} &       &       & 50.01  & 60.28  & 53.77  & 64.78  & 55.76  & 65.96  & 54.72  & 65.43  & 57.75  & 68.69  & 51.52  & 61.99  & 53.92  & 64.52  \\
    C & \textbf{\checkmark} & \textbf{\checkmark} & \textbf{\checkmark} &       & 55.12  & 66.58  & 55.45  & 66.85  & 57.22  & 68.85  & 55.36  & 65.71  & 57.46  & 68.49  & 53.68  & 64.58  & 55.72  & 66.84  \\
    \textbf{CTTA-T (ours)} & \textbf{\checkmark} & \textbf{\checkmark} & \textbf{\checkmark} & \textbf{\checkmark} & \textbf{57.85} & \textbf{69.72} & \textbf{57.99} & \textbf{69.76} & \textbf{58.00} & \textbf{69.80} & \textbf{60.84} & \textbf{72.08} & \textbf{60.92} & \textbf{72.18} & \textbf{60.86} & \textbf{71.96} & \textbf{59.41} & \textbf{70.92} \\
    \bottomrule
    \end{tabular}%
    }
    \caption{Ablation study of the four modules in our framework. \textbf{Bold}: the best results.}
    \label{tab:ablation}%
\end{table*}%

\section{Experiments}\label{sec:4}
\subsection{Experimental Settings}\label{4.1}
\textcolor{black}{\textbf{Tasks and Benchmarks.} We establish a text understanding CTTA benchmark with 11 sub-tasks spanning Robust QA \cite{ravichander-etal-2021-noiseqa} (speech, keyboard, translation noise), Cross-lingual QA \cite{artetxe2020cross} (Spanish, Chinese, Arabic), and Reading Comprehension \cite{fisch2019mrqa} (Search, Trivia). We construct short (7-task, Order 1-3) and long (11-task, Order 4-6) streams using one standard order (Order 1,4) and two randomized orders (Order 2,3,5,6) via Fisher–Yates shuffle. We also construct a sentiment analysis task stream (see details of the benchmark in App.~\ref{app:task}). 
}

\textbf{Baselines.}
\textit{Tent} \cite{wang2021tent} minimizes the prediction entropy.  
\textit{OIL} \cite{ye2022robust} employs a teacher-student framework and causal inference.  
\textit{CoTTA} \cite{wang2022continual} combines a teacher-student framework, mean predictions over augmentations.
\textit{SAR} \cite{niu2023towards} filters low-entropy samples and sharpens-aware minimization.
\textit{SoTTA} \cite{gong2023sotta} buffers class-balanced samples and sharpens-aware minimization.
\textit{REM} \cite{han2025ranked} enforces difficulty-aware adaptation via ranking and masked consistency.

\textcolor{black}{\textbf{Implementation Details.} Following prior work \cite{ye2022robust,su-etal-2023-beware}, we use XLM-RoBERTa-Base \cite{conneau2020unsupervised} and the robustness-tuned XLMR-Base-xTune \cite{zheng2021consistency} as backbones. Both are pre-trained on SQuAD \cite{rajpurkar-etal-2016-squad} using XTREME configurations \cite{hu2020xtreme}.
See more detailed \textbf{Implementation Details} and \textbf{Metrics} in App.~\ref{app:Implementation}.}

%

\subsection{Main Results}
Tab. \ref{tab: main_table} reports overall results across task streams (Orders 1 to 6), evaluated using exact match (EM) and F1 scores. Each score denotes the average performance after applying CTTA under the corresponding order.
Our method achieves SOTA performance with both backbones (i.e., the pre-trained model). 
Averaged over backbones, our method exhibits improvement over SAR (+3.96\% in EM, +4.19\% in F1) and REM (+12.31\% , +13.08\%), while outperforming the vanilla forward (no adaptation), Tent, OIL, CoTTA, and SoTTA. 
In addition, CTTA-T achieves more stable performance across orders for baselines (see App.~\ref{variance}).
\textcolor{black}{We further compare with strong baseline (see App.~\ref{app:Strong_Baseline}) and LLMs (see App.~\ref{app:zero-shot}).
Our method also achieves strong performance on the sentiment analysis task, as shown in Tab~\ref{tab:sentiment_analysis}. Our method outperforms SAR, REM, and vanilla forward, achieving accuracy improvements of 4.3\%, 6.81\%, and 2.48\%, respectively.}


\subsection{Ablation Studies}
We follow standard CTTA evaluation protocols \cite{wang2022continual,jiang2024pcotta} for ablation studies.
Tab.~\ref{tab:ablation} summarizes the contribution of each module.
Row~A (TS-CTTA, \S\ref{sec:3.1}) serves as the baseline.
Adding CDA enhances performance via dynamic cross-domain accumulation, enabling the teacher to gain domain awareness through IPCA-based semantic tracking.
RFS further improves stability and accuracy by enforcing prediction consistency and filtering uncertain guidance.
However, as shown in Row~C, prediction errors still accumulate and degrade performance over time.
SRT alleviates this issue by periodically restoring the teacher to the source state, reintroducing general knowledge.
Further analysis is provided in App.~\ref{app:Effect of Each Component}.
\begin{figure}[h]
    \centering
    \includegraphics[width=1\linewidth]{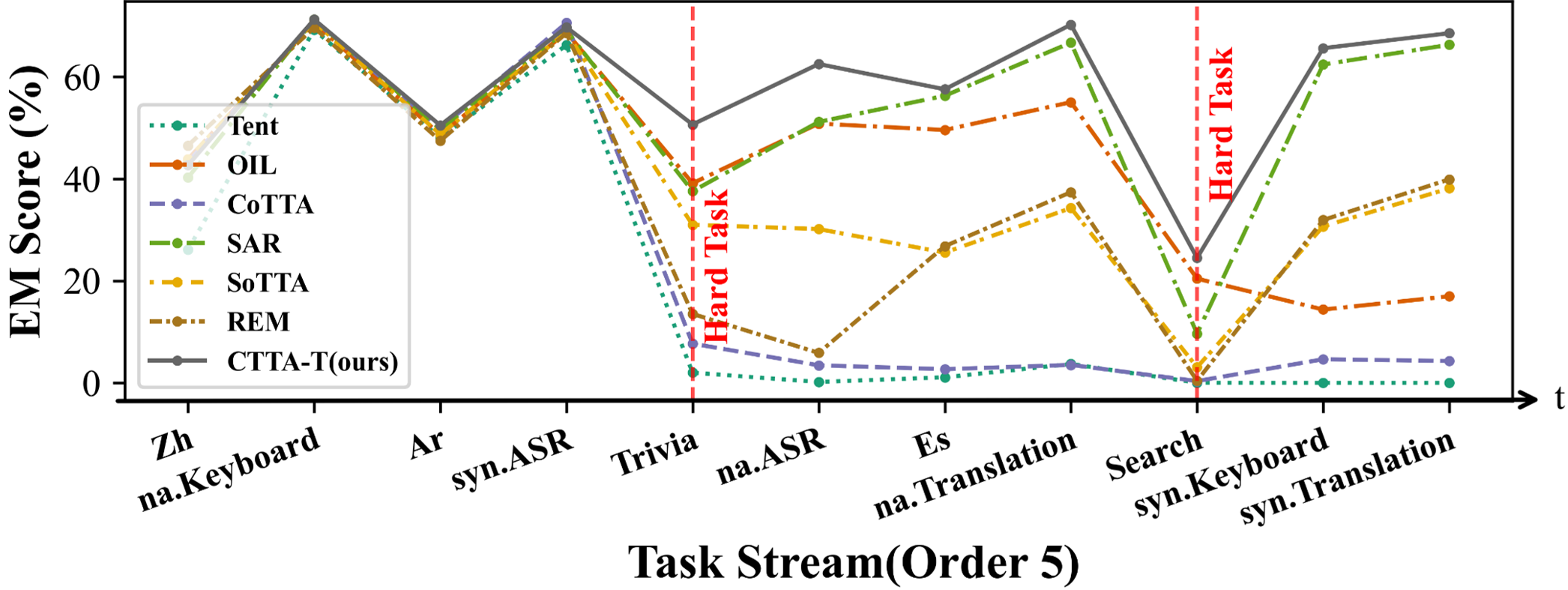}  
    \caption{
        Performance of ours and baselines on order 5. 
    }
    \label{fig:Model_collapse}
\end{figure}

\subsection{Analysis Studies}
\noindent \textbf{Analysis of Error Accumlation.}
To analyze error accumulation, we evaluate all methods on short and long task sequences, focusing on hard tasks (e.g., Search and Trivial), where error accumulation tends to be more severe.
As shown in Fig.~\ref{fig:Model_collapse}, all baselines suffered from error accumulation (i.e., collapse), while ours remained stable or showed only minimal decline. Highlighting ours effectively mitigates error accumulation.
Baseline collapse arises from uncontrolled error accumulation due to noisy pseudo-labels, though the underlying reasons differ across methods.
More analysis is in App.~\ref{app:futher Collapse}.

\begin{table}[htbp]
    \centering
    \caption{\textcolor{black}{Comparison of the cost of different methods.}}
    \label{tab:resource}%
    \resizebox{1\linewidth}{!}{
    \begin{tabular}{lccccc@{\hspace{8pt}}c}
    \toprule    
      & CTTA-T & REM & SoTTA & SAR & CoTTA & \textcolor{gray}{No adaptation} \\
    Backbone & Time (s) & Time (s) & Time (s) & Time (s) & Time (s) & \textcolor{gray}{Time (s)} \\
    \midrule
    Base & \textbf{0.043} & 0.061 & 0.051 & 0.044 & 0.196  & \textcolor{gray}{0.032}  \\
    xTune & \textbf{0.040} & 0.062 & 0.050 & 0.042 & 0.200 & \textcolor{gray}{0.032}  \\
    \bottomrule
    \end{tabular}
    }
\end{table}

\noindent \textcolor{black}{\textbf{Cost Comparison.}
We evaluate the method cost (per-sample adaptation plus inference time). Since baselines like Tent and OIL fail under continual shift (see Tab.~\ref{tab: main_table}), they are excluded. We compare with the strong baselines: CoTTA, SAR, SoTTA, and REM. Tab.~\ref{tab:resource} reports the average time across Orders. Compared with baselines, CTTA-T is the most efficient. Compared with the no-adaptation model (lowest bound), CTTA-T introduces only a small cost  (<35\%), which is acceptable.
The cost analyses of each module are in App.~\ref{app:cost_ablation}
}

\begin{figure}[h]
    \centering
    \includegraphics[width=1\columnwidth]{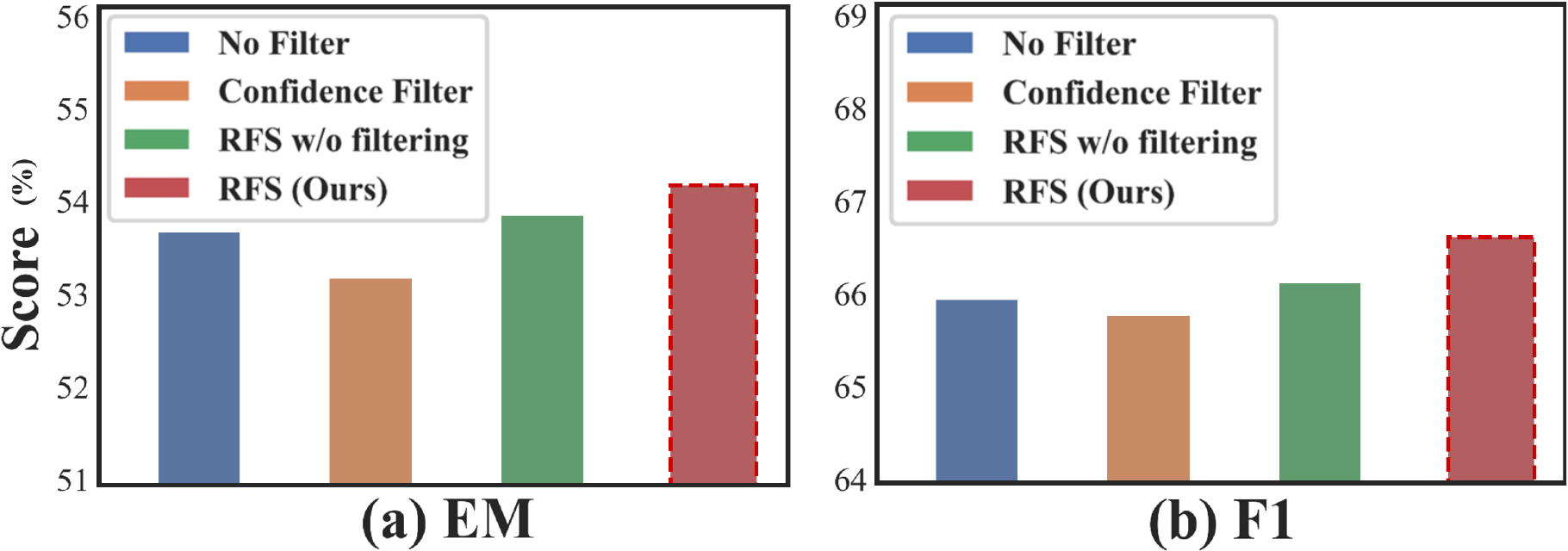}  
    \caption{
        Performance of different filtering strategies. 
    }
    \label{fig:SRF}
\end{figure}
\noindent{\textbf{Analysis of Filtering Strategies.}}
To assess the RFP's effectiveness, we compare it with other strategies: 1) no filter, 2) confidence filter, and 3) RFP w/o Filtering (disables RFP's filtering). 
As shown in Fig.~\ref{fig:SRF}, 
the RFP (red) outperformed all strategies, confirming its effectiveness in improving prediction by refining prediction and filtering noise.
Using only refining (green) ranks second, showing that refining improves prediction, while the confidence filter (orange) underperforms no filter (blue), indicating that only relying on prediction confidence is harmful in CTTA, where domain shifts cause overconfident wrong predictions.

\begin{figure}[h]
    \centering
    \includegraphics[width=1\columnwidth]{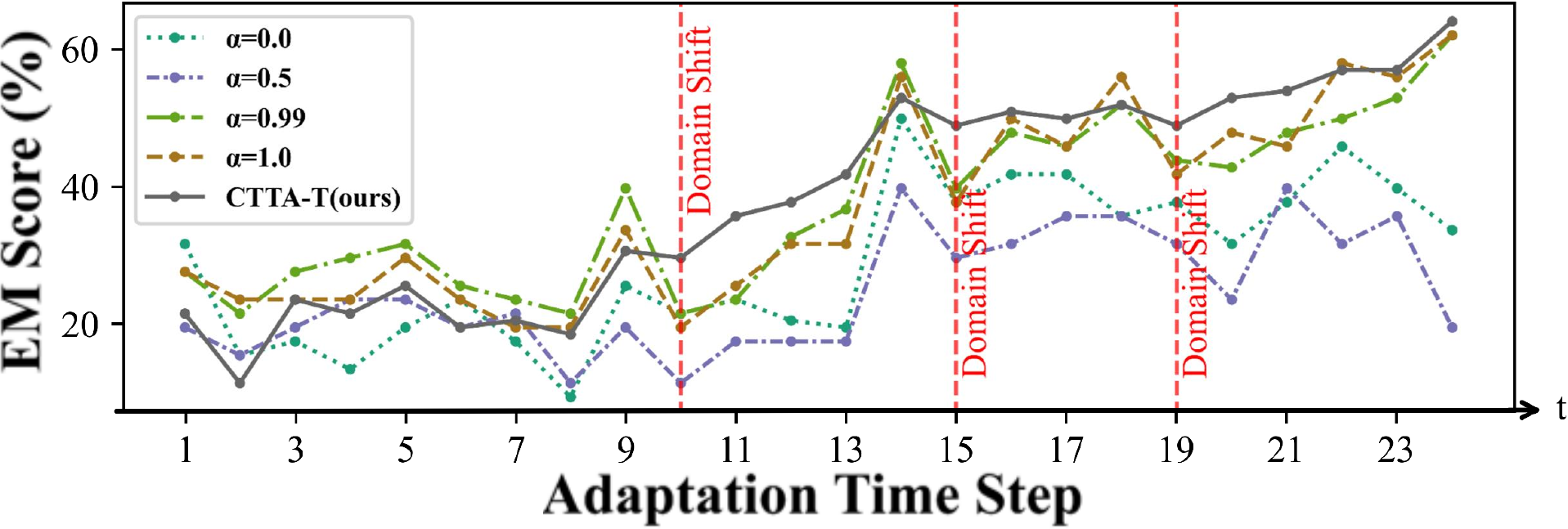}  
    \caption{
        Performance of our method and model using fixed weights under domain shift (report EM each step).
    }
    \label{fig:ipca}
\end{figure}
\noindent{\textbf{Analysis of CDA versus Fixed Teacher Updates.}}
To assess CDA under CTTA, we compared it with fixed weights ($\alpha{=}0/0.5/0.99/1.0$) on a stream with domain shifts at $t{=}10/15/19$.
As shown in Fig.~\ref{fig:ipca}, our method maintains or improves performance under shifts, while fixed-weight baselines drop sharply and recover slowly.
This advantage comes from CDA’s dynamic cross-domain accumulation, which balances domain-specific and domain-general semantic information, whereas fixed weights fail to adapt and cause instability.






\section{Conclusion}

We propose CTTA-T, a novel CTTA framework for text understanding. 
CTTA-T adopts a teacher-student architecture to support continuous test-time adaptation.  
A refine-then-filter module based on prediction consistency improves prediction reliability, while a domain-aware teacher dynamically accumulates cross-domain semantic knowledge and is partially restored to its source state to enhance generalization. Moreover, we introduce a benchmark for CTTA in text understanding. Experiments show that CTTA-T achieves strong performance.

\section*{Limitations}
\label{app:Limitation}
\textcolor{black}{Although CTTA-T shows strong and stable performance across diverse continual domain shift settings, there remain several natural limitations.
First, our benchmark primarily covers QA, reading comprehension, cross-lingual QA, and sentiment analysis. Notably, this evaluation scope is consistent with a large body of prior work in the CTTA and test-time adaptation literature, where methods are typically validated on a limited subset of representative tasks rather than the full spectrum of text understanding problems.
Extending CTTA to other text understanding tasks is our future work.
}

While we theoretically establish key properties (e.g., prediction consistency bound and domain-aware update rule), more general theoretical guarantees under arbitrary domain shifts remain open.

Importantly, these limitations do not undermine the practical effectiveness of our framework but instead suggest fruitful avenues for future research.

\bibliography{custom}

\begin{thebibliography}{42}
\providecommand{\natexlab}[1]{#1}

\bibitem[{Artetxe et~al.(2020)Artetxe, Ruder, Yogatama, and
  Liu}]{artetxe2020cross}
Mikel Artetxe, Sebastian Ruder, Dani Yogatama, and Liu. 2020.
\newblock On the cross-lingual transferability of monolingual representations.
\newblock In \emph{Proceedings of the 58th Annual Meeting of the Association
  for Computational Linguistics}, pages 4623--4637.

\bibitem[{Bartlett(1937)}]{bartlett1937properties}
Maurice~Stevenson Bartlett. 1937.
\newblock Properties of sufficiency and statistical tests.
\newblock \emph{Proceedings of the Royal Society of London. Series
  A-Mathematical and Physical Sciences}, 160(901):268--282.

\bibitem[{Boudiaf et~al.(2022)Boudiaf, Mueller, Ben~Ayed, and
  Bertinetto}]{boudiaf2022parameter}
Malik Boudiaf, Romain Mueller, Ismail Ben~Ayed, and Luca Bertinetto. 2022.
\newblock Parameter-free online test-time adaptation.
\newblock In \emph{Proceedings of the IEEE/CVF Conference on Computer Vision
  and Pattern Recognition}, pages 8344--8353.

\bibitem[{Cao et~al.(2020)Cao, Fang, Yu, and Zhou}]{cao2020unsupervised}
Yu~Cao, Meng Fang, Baosheng Yu, and Joey~Tianyi Zhou. 2020.
\newblock Unsupervised domain adaptation on reading comprehension.
\newblock In \emph{Proceedings of the AAAI Conference on Artificial
  Intelligence}, volume~34, pages 7480--7487.

\bibitem[{Chen et~al.(2019)Chen, Xie, Huang, Rong, Ding, Huang, Xu, and
  Huang}]{chen2019progressive}
Chaoqi Chen, Weiping Xie, Wenbing Huang, Yu~Rong, Xinghao Ding, Yue Huang,
  Tingyang Xu, and Junzhou Huang. 2019.
\newblock Progressive feature alignment for unsupervised domain adaptation.
\newblock In \emph{Proceedings of the IEEE/CVF conference on computer vision
  and pattern recognition}, pages 627--636.

\bibitem[{Chen et~al.(2022)Chen, Wang, Darrell, and
  Ebrahimi}]{chen2022contrastive}
Dian Chen, Dequan Wang, Trevor Darrell, and Sayna Ebrahimi. 2022.
\newblock Contrastive test-time adaptation.
\newblock In \emph{Proceedings of the IEEE/CVF Conference on Computer Vision
  and Pattern Recognition}, pages 295--305.

\bibitem[{Conneau et~al.(2020)Conneau, Khandelwal, Goyal, Chaudhary, Wenzek,
  Guzm{\'a}n, Grave, Ott, Zettlemoyer, and Stoyanov}]{conneau2020unsupervised}
Alexis Conneau, Kartikay Khandelwal, Naman Goyal, Vishrav Chaudhary, Guillaume
  Wenzek, Francisco Guzm{\'a}n, {\'E}douard Grave, Myle Ott, Luke Zettlemoyer,
  and Veselin Stoyanov. 2020.
\newblock Unsupervised cross-lingual representation learning at scale.
\newblock In \emph{Proceedings of the 58th Annual Meeting of the Association
  for Computational Linguistics}, pages 8440--8451.

\bibitem[{Devlin et~al.(2019)Devlin, Chang, Lee, and
  Toutanova}]{devlin-etal-2019-bert}
Jacob Devlin, Ming-Wei Chang, Kenton Lee, and Kristina Toutanova. 2019.
\newblock \href {https://doi.org/10.18653/v1/N19-1423} {{BERT}: Pre-training of
  deep bidirectional transformers for language understanding}.
\newblock In \emph{Proceedings of the 2019 Conference of the North {A}merican
  Chapter of the Association for Computational Linguistics: Human Language
  Technologies, Volume 1 (Long and Short Papers)}, pages 4171--4186,
  Minneapolis, Minnesota. Association for Computational Linguistics.

\bibitem[{Enevoldsen et~al.()Enevoldsen, Chung, Kerboua, Kardos, Mathur, Stap,
  Gala, Siblini, Krzemi{\'n}ski, Winata et~al.}]{enevoldsenmmteb}
Kenneth Enevoldsen, Isaac Chung, Imene Kerboua, M{\'a}rton Kardos, Ashwin
  Mathur, David Stap, Jay Gala, Wissam Siblini, Dominik Krzemi{\'n}ski,
  Genta~Indra Winata, and 1 others.
\newblock Mmteb: Massive multilingual text embedding benchmark.
\newblock In \emph{The Thirteenth International Conference on Learning
  Representations}.

\bibitem[{Fisch et~al.(2019)Fisch, Talmor, Jia, Seo, Choi, and
  Chen}]{fisch2019mrqa}
Adam Fisch, Alon Talmor, Robin Jia, Minjoon Seo, Eunsol Choi, and Danqi Chen.
  2019.
\newblock Mrqa 2019 shared task: Evaluating generalization in reading
  comprehension.
\newblock In \emph{Proceedings of the 2nd Workshop on Machine Reading for
  Question Answering}, pages 1--13.

\bibitem[{Gong et~al.(2023)Gong, Kim, Lee, Chottananurak, and
  Lee}]{gong2023sotta}
Taesik Gong, Yewon Kim, Taeckyung Lee, Sorn Chottananurak, and Sung-Ju Lee.
  2023.
\newblock Sotta: Robust test-time adaptation on noisy data streams.
\newblock \emph{Advances in Neural Information Processing Systems},
  36:14070--14093.

\bibitem[{Guo et~al.(2017)Guo, Pleiss, Sun, and
  Weinberger}]{guo2017calibration}
Chuan Guo, Geoff Pleiss, Yu~Sun, and Kilian~Q Weinberger. 2017.
\newblock On calibration of modern neural networks.
\newblock In \emph{International conference on machine learning}, pages
  1321--1330. PMLR.

\bibitem[{Han et~al.(2025)Han, Na, and Hwang}]{han2025ranked}
Jisu Han, Jaemin Na, and Wonjun Hwang. 2025.
\newblock \href {https://openreview.net/forum?id=lHaGLJ65J9} {Ranked entropy
  minimization for continual test-time adaptation}.
\newblock In \emph{Forty-second International Conference on Machine Learning}.

\bibitem[{Hardt and Sun(2024)}]{hardt2024testtime}
Moritz Hardt and Yu~Sun. 2024.
\newblock \href {https://openreview.net/forum?id=CNL2bku4ra} {Test-time
  training on nearest neighbors for large language models}.
\newblock In \emph{The Twelfth International Conference on Learning
  Representations}.

\bibitem[{Hu et~al.(2020)Hu, Ruder, Siddhant, Neubig, Firat, and
  Johnson}]{hu2020xtreme}
Junjie Hu, Sebastian Ruder, Aditya Siddhant, Graham Neubig, Orhan Firat, and
  Melvin Johnson. 2020.
\newblock Xtreme: A massively multilingual multi-task benchmark for evaluating
  cross-lingual generalisation.
\newblock In \emph{International conference on machine learning}, pages
  4411--4421. PMLR.

\bibitem[{Hu et~al.(2018)Hu, Peng, Huang, Qiu, Wei, and Zhou}]{ijcai2018p0570}
Minghao Hu, Yuxing Peng, Zhen Huang, Xipeng Qiu, Furu Wei, and Ming Zhou. 2018.
\newblock \href {https://doi.org/10.24963/ijcai.2018/570} {Reinforced mnemonic
  reader for machine reading comprehension}.
\newblock In \emph{Proceedings of the Twenty-Seventh International Joint
  Conference on Artificial Intelligence, {IJCAI-18}}, pages 4099--4106.
  International Joint Conferences on Artificial Intelligence Organization.

\bibitem[{Jeong et~al.(2023)Jeong, Baek, Cho, Hwang, and
  Park}]{jeong-etal-2023-test}
Soyeong Jeong, Jinheon Baek, Sukmin Cho, Sung Hwang, and Jong Park. 2023.
\newblock \href {https://doi.org/10.18653/v1/2023.findings-emnlp.1033}
  {Test-time self-adaptive small language models for question answering}.
\newblock In \emph{Findings of the Association for Computational Linguistics:
  EMNLP 2023}, pages 15459--15469, Singapore. Association for Computational
  Linguistics.

\bibitem[{Jiang et~al.(2024)Jiang, Zhou, Li, Zhao, Wang, Ma, Chang, Zhang, Lu
  et~al.}]{jiang2024pcotta}
Jincen Jiang, Qianyu Zhou, Yuhang Li, Xinkui Zhao, Meili Wang, Lizhuang Ma,
  Jian Chang, Jian Zhang, Xuequan Lu, and 1 others. 2024.
\newblock Pcotta: Continual test-time adaptation for multi-task point cloud
  understanding.
\newblock \emph{Advances in Neural Information Processing Systems},
  37:96229--96253.

\bibitem[{Karim et~al.(2023)Karim, Mithun, Rajvanshi, Chiu, Samarasekera, and
  Rahnavard}]{karim2023c}
Nazmul Karim, Niluthpol~Chowdhury Mithun, Abhinav Rajvanshi, Han-pang Chiu,
  Supun Samarasekera, and Nazanin Rahnavard. 2023.
\newblock C-sfda: A curriculum learning aided self-training framework for
  efficient source free domain adaptation.
\newblock In \emph{Proceedings of the IEEE/CVF conference on computer vision
  and pattern recognition}, pages 24120--24131.

\bibitem[{Kendall and Gal(2017)}]{kendall2017uncertainties}
Alex Kendall and Yarin Gal. 2017.
\newblock What uncertainties do we need in bayesian deep learning for computer
  vision?
\newblock \emph{Advances in neural information processing systems}, 30.

\bibitem[{Lee et~al.(2013)}]{lee2013pseudo}
Dong-Hyun Lee and 1 others. 2013.
\newblock Pseudo-label: The simple and efficient semi-supervised learning
  method for deep neural networks.
\newblock In \emph{Workshop on challenges in representation learning, ICML},
  volume~3, page 896. Atlanta.

\bibitem[{Lee et~al.()Lee, Jung, Lee, Park, Shin, Hwang, and Yoon}]{leeentropy}
Jonghyun Lee, Dahuin Jung, Saehyung Lee, Junsung Park, Juhyeon Shin, Uiwon
  Hwang, and Sungroh Yoon.
\newblock Entropy is not enough for test-time adaptation: From the perspective
  of disentangled factors.
\newblock In \emph{The Twelfth International Conference on Learning
  Representations}.

\bibitem[{Lin et~al.(2025)Lin, Wang, Liu, and Chen}]{lin-etal-2025-crossin}
Geyu Lin, Bin Wang, Zhengyuan Liu, and Nancy~F. Chen. 2025.
\newblock \href {https://aclanthology.org/2025.sumeval-2.2/} {{C}ross{I}n: An
  efficient instruction tuning approach for cross-lingual knowledge alignment}.
\newblock In \emph{Proceedings of the Second Workshop on Scaling Up
  Multilingual {\&} Multi-Cultural Evaluation}, pages 12--23, Abu Dhabi.
  Association for Computational Linguistics.

\bibitem[{Liu(2000)}]{liu2000statistical}
Y~Liu. 2000.
\newblock Statistical behavior of the eigenvalues of random matrices.
\newblock In \emph{Proceedings of the Mathematical Junior Seminar}.

\bibitem[{Loshchilov and Hutter(2019)}]{loshchilov2018decoupled}
Ilya Loshchilov and Frank Hutter. 2019.
\newblock \href {https://openreview.net/forum?id=Bkg6RiCqY7} {Decoupled weight
  decay regularization}.
\newblock In \emph{International Conference on Learning Representations}.

\bibitem[{Lyu et~al.(2024)Lyu, Du, Li, Zhao, Zhang, Liu, and
  Wang}]{lyu2024variational}
Fan Lyu, Kaile Du, Yuyang Li, Hanyu Zhao, Zhang Zhang, Guangcan Liu, and Liang
  Wang. 2024.
\newblock Variational continual test-time adaptation.
\newblock \emph{arXiv preprint arXiv:2402.08182}.

\bibitem[{Maas et~al.(2011)Maas, Daly, Pham, Huang, Ng, and
  Potts}]{maas-etal-2011-learning}
Andrew~L. Maas, Raymond~E. Daly, Peter~T. Pham, Dan Huang, Andrew~Y. Ng, and
  Christopher Potts. 2011.
\newblock \href {https://aclanthology.org/P11-1015/} {Learning word vectors for
  sentiment analysis}.
\newblock In \emph{Proceedings of the 49th Annual Meeting of the Association
  for Computational Linguistics: Human Language Technologies}, pages 142--150,
  Portland, Oregon, USA. Association for Computational Linguistics.

\bibitem[{Muennighoff et~al.(2023)Muennighoff, Tazi, Magne, and
  Reimers}]{muennighoff-etal-2023-mteb}
Niklas Muennighoff, Nouamane Tazi, Loic Magne, and Nils Reimers. 2023.
\newblock \href {https://doi.org/10.18653/v1/2023.eacl-main.148} {{MTEB}:
  Massive text embedding benchmark}.
\newblock In \emph{Proceedings of the 17th Conference of the European Chapter
  of the Association for Computational Linguistics}, pages 2014--2037,
  Dubrovnik, Croatia. Association for Computational Linguistics.

\bibitem[{Niu et~al.(2024)Niu, Miao, Chen, Wu, and Zhao}]{niu2024test}
Shuaicheng Niu, Chunyan Miao, Guohao Chen, Pengcheng Wu, and Peilin Zhao. 2024.
\newblock Test-time model adaptation with only forward passes.
\newblock In \emph{International Conference on Machine Learning}, pages
  38298--38315. PMLR.

\bibitem[{Niu et~al.(2023)Niu, Wu, Zhang, Wen, Chen, Zhao, and
  Tan}]{niu2023towards}
Shuaicheng Niu, Jiaxiang Wu, Yifan Zhang, Zhiquan Wen, Yaofo Chen, Peilin Zhao,
  and Mingkui Tan. 2023.
\newblock \href {https://openreview.net/forum?id=g2YraF75Tj} {Towards stable
  test-time adaptation in dynamic wild world}.
\newblock In \emph{The Eleventh International Conference on Learning
  Representations}.

\bibitem[{Rajpurkar et~al.(2016)Rajpurkar, Zhang, Lopyrev, and
  Liang}]{rajpurkar-etal-2016-squad}
Pranav Rajpurkar, Jian Zhang, Konstantin Lopyrev, and Percy Liang. 2016.
\newblock \href {https://doi.org/10.18653/v1/D16-1264} {{SQ}u{AD}: 100,000+
  questions for machine comprehension of text}.
\newblock In \emph{Proceedings of the 2016 Conference on Empirical Methods in
  Natural Language Processing}, pages 2383--2392, Austin, Texas. Association
  for Computational Linguistics.

\bibitem[{Ravichander et~al.(2021)Ravichander, Dalmia, Ryskina, Metze, Hovy,
  and Black}]{ravichander-etal-2021-noiseqa}
Abhilasha Ravichander, Siddharth Dalmia, Maria Ryskina, Florian Metze, Eduard
  Hovy, and Alan~W Black. 2021.
\newblock \href {https://doi.org/10.18653/v1/2021.eacl-main.259} {{N}oise{QA}:
  Challenge set evaluation for user-centric question answering}.
\newblock In \emph{Proceedings of the 16th Conference of the European Chapter
  of the Association for Computational Linguistics: Main Volume}, pages
  2976--2992, Online. Association for Computational Linguistics.

\bibitem[{Ross et~al.(2008)Ross, Lim, Lin, and Yang}]{2008Incremental}
David~A. Ross, Jongwoo Lim, Ruei~Sung Lin, and Ming~Hsuan Yang. 2008.
\newblock Incremental learning for robust visual tracking.
\newblock \emph{International Journal of Computer Vision}, 77(1-3):125--141.

\bibitem[{Socher et~al.(2013)Socher, Perelygin, Wu, Chuang, Manning, Ng, and
  Potts}]{socher-etal-2013-recursive}
Richard Socher, Alex Perelygin, Jean Wu, Jason Chuang, Christopher~D. Manning,
  Andrew Ng, and Christopher Potts. 2013.
\newblock \href {https://aclanthology.org/D13-1170/} {Recursive deep models for
  semantic compositionality over a sentiment treebank}.
\newblock In \emph{Proceedings of the 2013 Conference on Empirical Methods in
  Natural Language Processing}, pages 1631--1642, Seattle, Washington, USA.
  Association for Computational Linguistics.

\bibitem[{Su et~al.(2023)Su, Ji, Li, Ye, and Zhang}]{su-etal-2023-beware}
Yi~Su, Yixin Ji, Juntao Li, Hai Ye, and Min Zhang. 2023.
\newblock \href {https://doi.org/10.18653/v1/2023.emnlp-main.803} {Beware of
  model collapse! fast and stable test-time adaptation for robust question
  answering}.
\newblock In \emph{Proceedings of the 2023 Conference on Empirical Methods in
  Natural Language Processing}, pages 12998--13011, Singapore. Association for
  Computational Linguistics.

\bibitem[{Tarvainen and Valpola(2017)}]{tarvainen2017mean}
Antti Tarvainen and Harri Valpola. 2017.
\newblock Mean teachers are better role models: Weight-averaged consistency
  targets improve semi-supervised deep learning results.
\newblock \emph{Advances in neural information processing systems}, 30.

\bibitem[{Wang et~al.(2021)Wang, Shelhamer, Liu, Olshausen, and
  Darrell}]{wang2021tent}
Dequan Wang, Evan Shelhamer, Shaoteng Liu, Bruno Olshausen, and Trevor Darrell.
  2021.
\newblock \href {https://openreview.net/forum?id=uXl3bZLkr3c} {Tent: Fully
  test-time adaptation by entropy minimization}.
\newblock In \emph{International Conference on Learning Representations}.

\bibitem[{Wang et~al.(2022)Wang, Fink, Van~Gool, and Dai}]{wang2022continual}
Qin Wang, Olga Fink, Luc Van~Gool, and Dengxin Dai. 2022.
\newblock Continual test-time domain adaptation.
\newblock In \emph{Proceedings of the IEEE/CVF Conference on Computer Vision
  and Pattern Recognition}, pages 7201--7211.

\bibitem[{Wang et~al.(2024)Wang, Hong, Cheraghian, Rahman, Ahmedt-Aristizabal,
  Petersson, and Harandi}]{wang-2024-continual}
Yanshuo Wang, Jie Hong, Ali Cheraghian, Shafin Rahman, David
  Ahmedt-Aristizabal, Lars Petersson, and Mehrtash Harandi. 2024.
\newblock Continual test-time domain adaptation via dynamic sample selection.
\newblock In \emph{Proceedings of the IEEE/CVF Winter Conference on
  Applications of Computer Vision}, pages 1701--1710.

\bibitem[{Ye et~al.(2022)Ye, Ding, Li, and Ng}]{ye2022robust}
Hai Ye, Yuyang Ding, Juntao Li, and Hwee~Tou Ng. 2022.
\newblock Robust question answering against distribution shifts with test-time
  adaption: An empirical study.
\newblock In \emph{Findings of the Association for Computational Linguistics:
  EMNLP 2022}, pages 6179--6192.

\bibitem[{Yue et~al.(2022)Yue, Zeng, Kou, Shang, and Wang}]{yue2022contrastive}
Zhenrui Yue, Huimin Zeng, Ziyi Kou, Lanyu Shang, and Dong Wang. 2022.
\newblock Contrastive domain adaptation for early misinformation detection: A
  case study on covid-19.
\newblock In \emph{Proceedings of the 31st ACM international conference on
  information \& knowledge management}, pages 2423--2433.

\bibitem[{Zheng et~al.(2021)Zheng, Dong, Huang, Wang, Chi, Singhal, Che, Liu,
  Song, and Wei}]{zheng2021consistency}
Bo~Zheng, Li~Dong, Shaohan Huang, Wenhui Wang, Zewen Chi, Saksham Singhal,
  Wanxiang Che, Ting Liu, Xia Song, and Furu Wei. 2021.
\newblock Consistency regularization for cross-lingual fine-tuning.
\newblock In \emph{Proceedings of the 59th Annual Meeting of the Association
  for Computational Linguistics and the 11th International Joint Conference on
  Natural Language Processing (Volume 1: Long Papers)}, pages 3403--3417.

\end{thebibliography}

\appendix
\appendix
\section{Problem Definition}
\label{app:Problem Definition}
\begin{figure*}[htbp]
    \centering
    \includegraphics[width=0.85\textwidth]{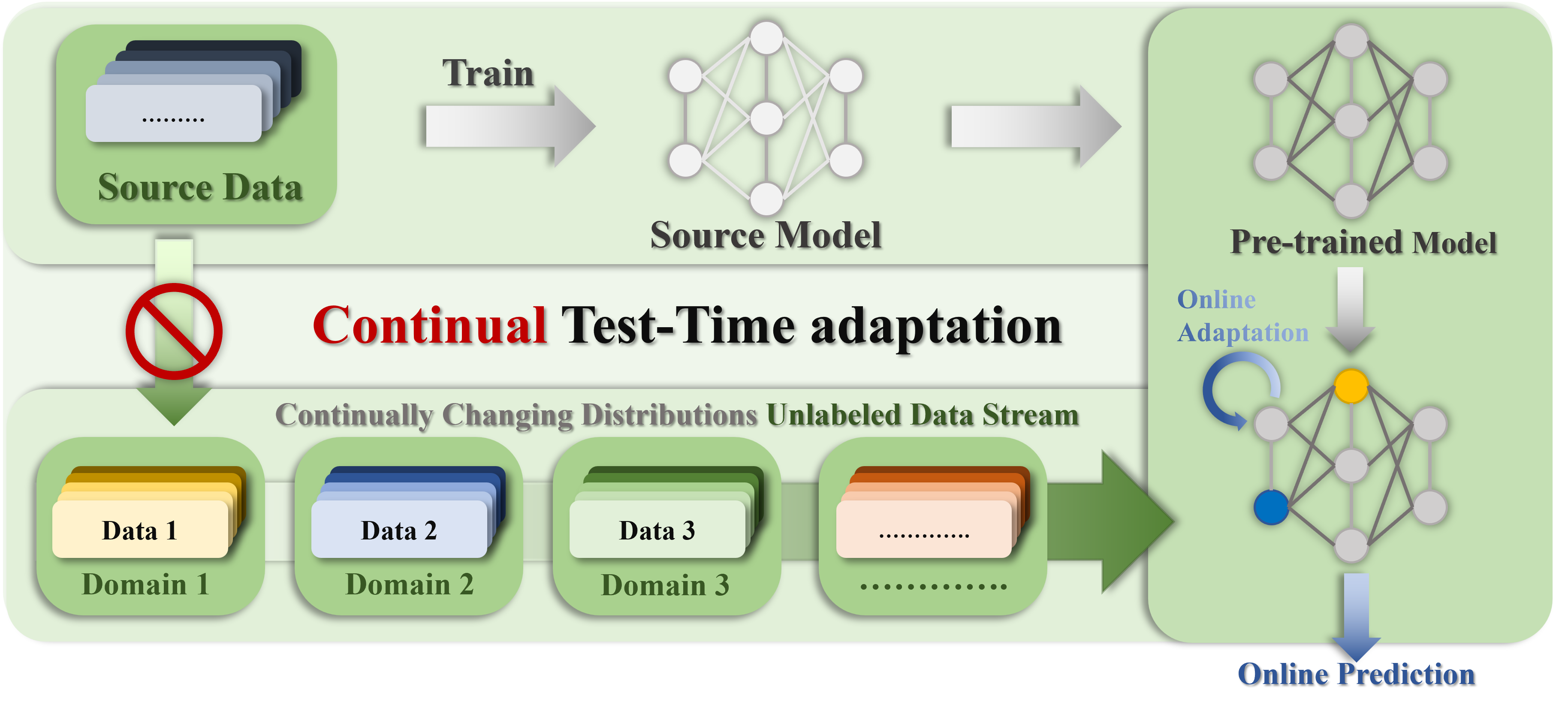}  
    \caption{
        Illustrating the setting of Continual Test-Time Adaptation (CTTA).
        To better reflect real-world environment, the target test data arrives sequentially, with its domain shifting continuously over time.
        At each time step, the model only accesses the current data.
        A model pre-trained on the source domain is used to perform online prediction and adaptation to the current target data, without access to any source domain data.
    }
    \label{fig:CTTA_setting}
\end{figure*}

Let the data distribution of the source domain be denoted as \(P_s\), and the data distribution of the target domain vary over time as \(P_t\), where \(t\) indicates the time step. Given an initial model \(\theta_0\), it is pre-trained on the source domain dataset \(\{X^s, Y^s\} \sim P_s\). At each time step \(t\), the model receives an unlabeled test sample \(x_t^T \in X_t^T\) from the target domain and produces a prediction \(y_t^T \in Y_t^T\), where \(\{X_t^T, Y_t^T\} \sim P_t\). The model then immediately updates its parameters \(\theta_t\) using the pseudo-label \(y_t^T\). 
It is important to note that during the adaptation process, the model has no access to the source training data \(X^s\), past test samples \(X_{<t}^T\), or future test samples \(X_{>t}^T\); it can only utilize the current test instance \(X_t^T\) for online adaptation. Our objective is to enable the model to continuously adapt to the evolving target distribution during test time, thereby enhancing its prediction performance in the target domain. The overall procedure of the CTTA setting is illustrated in Fig.~\ref{fig:CTTA_setting}.

\section{Theoretical derivation of the CDA module}
\label{app:DWU}
\subsection{The derivation process of IPCA}
\label{app:b.1}
Let the input embedding at time step \( t \) be denoted by \( X_{t} \in \mathbb{R}^{n \times d} \), where \( n \) represents the number of samples and \( d \) is the embedding dimension. The symmetric covariance matrix \( C_{t} \) and the mean \( \mu_{t} \) of the historical data are defined as
\begin{equation}
\begin{aligned}
C_t &= \frac{1}{N_t}\sum_{i=1}^t (X_i - \mu_t)^\top (X_i - \mu_t), \\
\mu_t &= \frac{1}{N_t}\sum_{i=1}^t \sum_{j=1}^n X_i^{(j)}.
\end{aligned}
\end{equation}
where \( N_{\mathrm{t}} \) denotes the number of historical samples. SVD is applied to \( C_{t} = V_{t} \Lambda_{t} V_{t}^{\mathsf{T}} \), extracting the principal component space \( V_{t} \in \mathbb{R}^{d \times k} \). When the new samples \( X_{t+1} \) arrives, the statistics are recursively updated using IPCA: 
\begin{equation}\label{eq:11}
\begin{aligned}
C_{t+1} \approx {} & 
\frac{N_tC_t+nC_\mathrm{new}}
     {N_t+n}  \\
&+ \frac{N_tn}{N_t+n}\,
   \frac{\left(\mu_t-\mu_\mathrm{new}\right)
         \left(\mu_t-\mu_\mathrm{new}\right)^\top}{N_t+n}, \\
\mu_{t+1} = {} &
\frac{N_t\mu_t+n\mu_\mathrm{new}}{N_t+n}.
\end{aligned}
\end{equation}
where \(C_{\mathrm{new}}=\frac{1}{n}(X_{t+1}-\mu_{\mathrm{new}})^{\top}(X_{t+1}-\mu_{\mathrm{new}})\) denotes the covariance matrix after the new sample $X_{t+1}$ is added. 
\(\mu_{\mathrm{new}}=\frac{1}{n}\sum_{j=1}^{n}X_{t+1}^{(j)}\) denotes the mean of the embedding layer after the new sample $X_{t+1}$ is added.

\subsection{Proof of Domain Semantic Distance between Historical and Current Domain}

\label{app:b.2}
Let $\Delta V = V_{t+1} - V_{t}$ denote the change in the principal directions after incorporating a new batch of data into the principal component space. The principal component space of the historical data are denoted as $V = [v_{t,1}, v_{t,2}, \ldots, v_{t,i}, \ldots, v_{t,k}]$, while after adding the new data, the updated principal component space are given by $V_{t+1} =  [v_{t+1,1}, v_{t+1,2}, \ldots, v_{t+1,i}, \ldots, v_{t+1,k}]$, where $v_{t+1,i}$ represents the $i$-th principal component direction.
We define the principal component similarity as the average cosine similarity between the old and new principal component spaces:
\begin{equation}
similarity=\frac{1}{k}\sum_{i=1}^{k}\cos\theta_{i}=\frac{1}{k}\sum_{i=1}^{k}(v_{t,i}^{\top}\cdot v_{t+1,i}),
\end{equation} 

where $\theta_{i}$ denotes the angle between $v_{t,i}$ and $v_{t+1,i}$. Since both $v_{t,i}$ and $v_{t+1,i}$ are unit vectors, the cosine similarity can be expressed as $\cos\theta_{i}=v_{t,i}^{\mathsf{T}}\cdot v_{t+1,i}.$ Given that the new batch is small relative to the historical data, the change in principal directions is slight, implying that $\theta_{i}$ is also small. Thus, using a Taylor expansion, we have the approximation
\begin{equation}
\cos\theta_i\approx1-\frac{\theta_i^2}{2}
\end{equation}

Let $\Delta v_{i}$ denote the perturbation of the $i$-th principal component direction, so that $v_{t+1,i} = v_{t,i} + \Delta v_{i}.$ Since both $v_{t,i}$ and $v_{t+1,i}$ are unit vectors, it follows that
\begin{equation}
\begin{aligned}
\|v_{t+1,i}\|^2 
  &= 1 = \|v_{t,i}+\Delta v_i\|^2 \\
  &= 1 + 2v_{t,i}^\top \Delta v_i 
     + \|\Delta v_i\|^2 .
\end{aligned}
\end{equation}

which leads to $v_{t,i}^\top\Delta v_{i} = -\frac{\parallel \Delta v_{i} \parallel^{2}}{2}.$ 
The cosine similarity between $v_{t,i}$ and $v_{t+1,i}$ can thus be expanded as
\begin{equation}
\cos\theta_i=v_{t,i}^\top\cdot v_{t+1,i}=v_{t,i}^\top(v_{t,i}+\Delta v_i)=1+v_{t,i}^\top\Delta v_i,
\end{equation}

hence: 
\begin{equation}
\begin{aligned}
1 + v_{t,i}^\top \Delta v_i &= 1 - \frac{\theta_i^2}{2} \Longrightarrow \\
 \quad 
v_{t,i}^\top \Delta v_i &= -\frac{\|\Delta v_i\|^2}{2} \approx -\frac{\theta_i^2}{2}.
\end{aligned}
\end{equation}

Consequently, the angle $\theta_{i}$ can be approximated as $\theta_{i} \approx \parallel \Delta v_{i} \parallel$. By substituting this approximation into the Taylor expansion of the cosine similarity, we obtain:
\begin{equation}
\cos\theta_i\approx1-\frac{\parallel\Delta v_i\parallel^2}{2}.
\end{equation}

Substituting this expression into the definition of principal component similarity yields:
\begin{align}
similarity&=\frac{1}{k}\sum_{i=1}^{k}\cos\theta_{i}\\
&\approx\frac{1}{k}\sum_{i=1}^{k}\left(1-\frac{\parallel\Delta v_{i}\parallel^{2}}{2}\right)\\
&=1-\frac{1}{2k}\sum_{i=1}^{k}\parallel\Delta v_{i}\parallel^{2}.
\end{align}

Let $\Delta V = [\Delta v_{1}, \Delta v_{2}, \ldots, \Delta v_{k}]$ denote the perturbation matrix, whose Frobenius norm satisfies
\begin{equation}
\parallel\Delta V\parallel_F^2=\sum_{i=1}^k\parallel\Delta v_i\parallel^2,
\end{equation}

where $\parallel \cdot \parallel_F$ denotes the Frobenius norm. Accordingly, the principal component similarity can be approximated as: 
\begin{equation}\label{eq:23}
similarity\approx1-\frac{1}{2k}\parallel\Delta V\parallel_F^2.
\end{equation}
To intuitively capture domain shifts, we use the distance as a proxy for domain similarity and simplify Eq. \ref{eq:23}:
\begin{equation}
distance = 2k\cdot(1-similarity)=\parallel\Delta V\parallel_F^2
\end{equation}
\subsection{Proof for the CDA Module's Robustness to Domain Shifts}
\label{app:b.3}
\textit{\textbf{Lemma 1.} \( \Delta V \) lies in the orthogonal complement of \( V_t \), it sensitively captures subtle shifts, enabling accurate quantification}.

\textbf{Proof.} Since both $V_{t}$ and $V_{t+1}$ are orthogonal matrices, they satisfy the orthogonality condition $V_{t+1}^{\mathrm{T}}V_{t+1}=I$, where $I$ denotes the identity matrix. 
By substituting $V_{t+1}=V_{t}+\Delta V$ into the orthogonality condition, we obtain:
\begin{equation}
\begin{aligned}
(V_t+\Delta V)^\top(V_t+\Delta V) &= I \\
V_t^\top V_t + V_t^\top\Delta V + \Delta V^\top V_t + \Delta V^\top\Delta V &=I.
\end{aligned}
\end{equation}

Given that $V_{t}^{\mathrm{T}}V_{t}=I$ and neglecting the second-order infinitesimal term $\Delta V^{\top}\Delta V \approx 0$, it follows that: $V_{t}^{\mathrm{T}}\Delta V+\Delta V^{\mathrm{T}}V_{t}=0.$ Rearranging the terms, we conclude that $\Delta V$ resides in the orthogonal complement space of $V$.

When the shift $\Delta V$ ($\Delta V= V_{t+1} - V_t$) between the current domain $V_t$ and the newly arrived domain $V_{t+1}$  is too subtle, IPCA can still effectively capture it. This is because the subtle shift $\Delta V$ lies along directions orthogonal to the current principal component space $V_t$, so it lies along directions that are previously unseen with the $V_t$. Then, these previously unseen directions will not be suppressed by $V_t$, thus, even when $\Delta V$ is very subtle, it will still produce an observable effect with $V_t$, allowing IPCA to sensitively detect such subtle shifts $\Delta V$.

\noindent\textcolor{black}{\textit{\textbf{Lemma 2.}} The domain shifts occurring in non-primary directions will be pushed into the primary components.}

\textcolor{black}{
\textbf{Proof.} When a new domain arrives, it changes the energy distribution of the global covariance matrix, which in turn causes a rotation of the principal component subspace. This rotation allows directions that originally had low variance (and thus were not part of the primary components) to become part of the updated principal component space $V_{t+1}$. 
As a result, even if the domain shift occurs mainly along non-primary component directions, IPCA can still capture these changes by gradually incorporating them into the evolving principal subspace.}

\textcolor{black}{
According to Eq.~\ref{eq:11} in the paper, the covariance matrix $C$ is updated incrementally:
$$C_{t+1} \approx \frac{N_t C_t + n C_{new}}{N_t + n}.$$
Assume that the current domain shift mainly occurs in non-primary directions of $V_t$ (i.e., those corresponding to very small eigenvalues in $C_t$, representing the noise subspace).} 

\textcolor{black}{
When the new batch of data (containing the domain shift) arrives, its covariance matrix $C_{\text{new}}$ contains significant variance along this "non-primary direction". Through the weighted averaging update above, the corresponding entries in $C_{t+1}$ will increase significantly. When we perform SVD on the updated $C_{t+1}$ to obtain $V_{t+1}$, the increased energy (eigenvalue) in this direction leads to reordering among the original principal components, causing a rotation of the eigenvector basis.}  

\textcolor{black}{
The direction that previously belonged to the noise subspace becomes the new $k$-th principal component. Since $\Delta V = V_{t+1} - V_t$, any such rotation of $V_{t+1}$ (i.e., when the new and old principal component spaces no longer coincide) will yield a significant value in $\Vert\Delta V\Vert_F^2$.}  

\textcolor{black}{
Therefore, when a shift carries informative variation, even if it initially lies in a non–principal component direction, it will typically enter the principal component space $V_{t+1}$ through the cumulative update of the covariance matrix. As a result, in most cases, IPCA is capable of sensing shifts that occur outside the original primary component space.}

\subsection{Derivation of the Relationship between CDA Weight $\alpha$ and Domain Semantic Distance}
\label{app:b.4}
We construct the following regularization function $\mathcal{F}$:
\begin{equation}
\begin{aligned}
\mathcal{F}(\theta_{t+1}^T) 
&= \parallel\Delta V\parallel_F^2 \, (\theta_{t+1}^T - \theta_{t+1}^S)^2 \\
&\quad + (1-\parallel\Delta V\parallel_F^2) \, (\theta_{t+1}^T - \theta_t^T)^2
\end{aligned}
\end{equation}

By differentiating $\mathcal{F}$ concerning $\theta_{t+1}^{T}$ and setting the derivative to zero, we obtain the optimal update direction for the teacher parameters:
\begin{equation}
\begin{aligned}
\frac{\partial \mathcal{F}}{\partial \theta_{t+1}^T} 
&= 2 \, \parallel\Delta V\parallel_F^2 (\theta_{t+1}^T - \theta_{t+1}^S) \\
&\quad + 2 \, (1-\parallel\Delta V\parallel_F^2) (\theta_{t+1}^T - \theta_t^T) \\
&= 0
\end{aligned}
\end{equation}

Rearranging the terms, we have:
\begin{equation}
\begin{aligned}
\theta_{t+1}^{T} 
&= \frac{(1-\parallel\Delta V\parallel_F^2)\theta_{t}^{T} 
       + \parallel\Delta V\parallel_F^2 \theta_{t+1}^{S}}
       {(1-\parallel\Delta V\parallel_F^2) + \parallel\Delta V\parallel_F^2} \\
&= (1-\parallel\Delta V\parallel_F^2)\theta_t^T 
   + \parallel\Delta V\parallel_F^2 \theta_{t+1}^S.
\end{aligned}
\end{equation}

Meanwhile, the standard EMA update formula is given by:
\begin{equation}
\theta_{t+1}^T=\alpha\theta_t^T+(1-\alpha)\theta_{t+1}^S.
\end{equation}

By applying the method of undetermined coefficients, it follows that:
\begin{equation}
    \alpha=1-\parallel\Delta V\parallel_F^2=1-distance.
\end{equation}
Since $\alpha$ varies with time step $t$, it is appropriate to define $\alpha$ as $\alpha_t$. Due to the model’s high sensitivity to parameter changes, we map \( \alpha_t \) to the range \([ \alpha_{\text{bound}}, 1 ]\).
\begin{table*}[htbp]
\centering
\caption{Task sequences (Orders 1–6) in the CTTA benchmark.
Orders 1–3 are short sequences comprising 7 tasks, and Orders 4–6 are long sequences comprising 11 tasks.}
\resizebox{\textwidth}{!}{
    \begin{tabular}{cl}
    \toprule
    \textbf{Order} & \multicolumn{1}{c}{\textbf{Task Sequence}} \\
        \midrule
        1 & syn.Keyboard  $\rightarrow$ syn.ASR  $\rightarrow$ syn.Translation  $\rightarrow$ Search  $\rightarrow$ Zh  $\rightarrow$ Ar  $\rightarrow$ Es \\
        2 & syn.ASR  $\rightarrow$ Es  $\rightarrow$ Zh  $\rightarrow$ syn.Keyboard  $\rightarrow$ Search  $\rightarrow$ syn.Translation  $\rightarrow$ Ar \\
        3 & Es  $\rightarrow$ syn.ASR  $\rightarrow$ syn.Keyboard  $\rightarrow$ Zh  $\rightarrow$ syn.Translation  $\rightarrow$ Search  $\rightarrow$ Ar \\
        \midrule
        4 & syn.Keyboard  $\rightarrow$ syn.ASR  $\rightarrow$ syn.Translation  $\rightarrow$ na.Keyboard  $\rightarrow$ na.ASR  $\rightarrow$ na.Translation  $\rightarrow$ Search  $\rightarrow$ Trivia  $\rightarrow$ Zh  $\rightarrow$ Ar  $\rightarrow$ Es \\
        5 & Zh  $\rightarrow$ na.Keyboard  $\rightarrow$ Ar  $\rightarrow$ syn.ASR  $\rightarrow$ Trivia  $\rightarrow$ na.ASR  $\rightarrow$ Es  $\rightarrow$ na.Translation  $\rightarrow$ Search  $\rightarrow$ syn.Keyboard  $\rightarrow$ syn.Translation \\
        6 & syn.Translation  $\rightarrow$ Zh  $\rightarrow$ Trivia  $\rightarrow$ Ar  $\rightarrow$ na.ASR  $\rightarrow$ Search  $\rightarrow$ na.Translation  $\rightarrow$ syn.Keyboard  $\rightarrow$ syn.ASR  $\rightarrow$ Es  $\rightarrow$ na.Keyboard \\
        \bottomrule
    \end{tabular}
    \label{tab:task_sequences}
}
\end{table*}

\section{Theoretical derivation of the RFP module}
\label{app:SRF}

\subsection{The Proof of the Upper Bound of the Consistency Score}
\label{app:c.1}
Given a classification model with $C$ output classes and a stochastic dropout mechanism, we denote by $p_i\in\mathbb{R}^C$ the predicted softmax probability vector of a single input sample after the $i$-th stochastic forward pass (with dropout enabled). Repeating this process for $N$ stochastic passes yields a prediction consistency matrix $P$ (The main text uses $P_C$):
\begin{equation}P=
\begin{bmatrix}
p_1 \\
p_2 \\
\vdots \\
p_N
\end{bmatrix}\in\mathbb{R}^{N\times C}.\end{equation}
Each row of P corresponds to the model's probabilistic prediction under a specific dropout mask, and collectively, $P$ encodes the model’s belief distribution consistency under randomness.

We focus on high-noise or ambiguous samples, which are typically difficult to classify due to a lack of discriminative information. Models tend to produce less confident and more uniform predictions for such inputs. 
For a high-noise sample, each row vector $p_i\in\mathbb{R}^C$  in $P$ satisfies: $\forall i\in\{1,\ldots,N\},\quad\mathbf{p}_i\approx\mathbf{u}_C$, where $\mathbf{u}_{C}=\left[\frac{1}{C},\ldots,\frac{1}{C}\right]\in\mathbb{R}^{C}$ denotes the uniform distribution over classes.

We perform singular value decomposition (SVD) on $P$:
\begin{equation}
P=U\Sigma V^\top,
\end{equation}
where $\mathbf{U}\in\mathbb{R}^{N\times N}$ is the matrix of left singular vectors, $\Sigma=\mathrm{diag}(\sigma_1,\sigma_2,\ldots,\sigma_r)\in\mathbb{R}^{N\times C}$
where $U\in\mathbb{R}^{N\times N}$ is the diagonal matrix of singular values $\sigma_1\geq\sigma_2\geq\cdots\geq\sigma_r>0$, $V\in\mathbb{R}^{C\times C}$ is the matrix of right singular vectors, which corresponds to directions in the class space.
Since $V^\top$ spans the class space, it captures how predictions distribute and align along different class directions. Specifically, each row $v_j\in\mathbb{R}^C~in~V$ represents a class-wise projection of a principal direction in the model's output distribution.

We define a class-level prediction consistency vector $\vec{s}\in\mathbb{R}^C$ , where each element $s_j$ reflects how strongly the principal components align with class $j$. The score is computed as a weighted sum over principal components projected onto the class $j$:
\begin{equation}
\mathbf{s}_j=\sum_{k=1}^r\sigma_k\cdot v_{k,j},
\end{equation}
where $v_{k,j}$ is the $j$-th entry of the $k$-th right singular vector (i.e., the projection of the $k$-th component of the class $j$), $\sigma_{k}$ is the singular value corresponding to the $k$-th component, which serves as its importance weight. 
A peaky consistency score $\mathbf{s}_j$ (i.e., one class dominates) indicates that the principal directions of the model's predictions are aligned toward a particular class, reflecting high agreement across dropout samples.
A flat $\mathbf{s}_j$ suggests that the model's predictions fluctuate in multiple directions and no class dominates, indicating low consistency and potentially high noise.

We define a consistency score for each test sample:
\begin{equation}\mathbf{s}_{\max}=\max_{j\in\{1,\ldots,\mathcal{C}\}}\mathbf{s}_{j},
\end{equation}
We set a filtering threshold $\tau$ and \textbf{discard samples whose $\mathbf{s}_{\text{max}} < \tau$}. Below, we provide theoretical justification for this filtering rule by proving that noisy samples are likely to exhibit lower $\mathbf{s}_{\text{max}}$ values.

Define the \textbf{average prediction} vector:
\begin{equation}
\bar{p} = \frac{1}{N} \sum_{i=1}^N p_i.
\end{equation}
Then, the consistency score satisfies:
\begin{equation}
\mathbf{s}_{\text{max}} \leq \max_j \bar{p}_j + \epsilon,
\end{equation}
where $\epsilon$ is a function of the variance among the $p_i$ and alignment between singular vectors and class axes. 
When $p_i$ is stable and aligned (low-noise), singular vectors align with dominant class $j^*$, and softmax amplifies this; when $p_i$ is diverse (high-noise), $\bar{p}$ becomes flat and singular directions point to mixed class subspaces; this reduces the $\vec{s}_j$ value for any $j$, leading to lower $\mathbf{s}_{\text{max}}$.
In conclusion, when the sample $x_t$ is dominated by data noise,  the consistency matrix \( P \) exhibits random behavior and causes a drop in \( \mathbf{s}_{\text{max}}\). 
Therefore, setting a threshold $\tau$ to filter out the lower $\mathbf{s}_{\text{max}}$ enables effective masking of samples with high AU.

\textcolor{black}{
\subsection{The Tightness Analysis of the Upper Bound of the Consistency Score}
As discussed in Sec.~\ref{sec:3.2}, the bound $s_{\max} \le \max_j p_j + \epsilon$ is derived specifically to account for the inherent noise in the data. 
Our theoretical derivation shows that the bound remains tight across all noise conditions. }

\textcolor{black}{
Under both high-noise and low-noise conditions (the two extremes), $s_{\max} \approx \max_j p_j + \epsilon$. In the medium-noise regime, $\max_j p_j + \epsilon$ is slightly larger than $s_{\max}$, ensuring that the bound closely approximates $s_{\max}$ across all noise conditions. }

\noindent\textcolor{black}{\textbf{Detailed derivation:}
We have the consistency matrix $P = U \Sigma V^\top$, the consistency vector $s = \sum_{k=1}^{r} \sigma_k v_k$, and the average prediction vector $\bar{p} = \frac{1}{N} \sum_{i=1}^{N} p_i = \frac{1}{N} P^\top 1_N$. For each class $j$, 
\begin{equation}
    s_j - \bar p_j = \sum_{k=1}^{r} \sigma_k v_{k,j} \left(1 - \frac{1}{N} u_k^\top 1_N\right),
\end{equation}
so that
\begin{equation}
\vert s_j - \bar p_j\vert \le \sum_{k=1}^{r} \sigma_k \vert v_{k,j}\vert \, \big\vert 1 - \frac{1}{N} u_k^\top 1_N \big\vert \le \epsilon,
\end{equation}
where $\epsilon = \max_j \sum_{k=1}^{r} \sigma_k \vert v_{k,j}\vert \, \big\vert 1 - \frac{1}{N} u_k^\top 1_N \big\vert.$
Thus, for the maximal class $j^\ast$:
\begin{equation}\label{eq:app_1}
\begin{aligned}
s_{\max} &= s_{j^\ast} \\
&= \bar p_{j^\ast} + \sum_{k=1}^{r} \sigma_k \vert v_{k,j^\ast}\vert \big\vert 1 - \frac{1}{N} u_k^\top 1_N \big\vert \\
&\le \bar p_{\max} + \epsilon.
\end{aligned}
\end{equation}
}
\textcolor{black}{
Noise conditions case analysis:
\begin{itemize}
    \item Low-noise samples: $p_i$ are highly aligned with a dominant class $j^\ast$. Then $\bar p_{\max} \approx 1$, $s_{\max} \approx \sqrt{N}$, and $\epsilon \approx (\sqrt{N}-1) \bar p_{\max}$. So, $\bar p_{\max} + \epsilon \approx s_{\max}$. The bound is tight.
    \textcolor{black}{
    \item High-noise samples: $p_i$ are nearly uniform. Then $\bar p_{\max} \approx 1/C$, $s_{\max} \approx \sqrt{N}/C$, and $\epsilon \approx (\sqrt{N}-1)/C$. So, $\bar p_{\max} + \epsilon \approx s_{\max}$. The bound is tight.}
    \textcolor{black}{
    \item Intermediate-noise samples: Assume $p_i$ are partially concentrated, i.e., neither fully aligned nor fully uniform. The $\epsilon$ is very close to the sum for the maximal class $j^\ast$, because $j^\ast$ naturally selects the class whose singular vectors align most with the principal directions (as shown in Eq.~\ref{eq:app_1}).
    }
\end{itemize}
Therefore, even for intermediate-noise samples, the alignment term is rigorously upper-bounded by $\epsilon$, guaranteeing the tightness of the bound.
}

\section{Implementation Details and Metrics}
\label{app:Implementation}
\noindent\textbf{Implementation Details.}
In the QA task, we employ XLM-RoBERTa-Base \cite{conneau2020unsupervised} and XLMR-Base-xTune as backbone models, following configurations used in prior studies\cite{ye2022robust}\cite{su-etal-2023-beware}. 
The latter uses xTune\cite{zheng2021consistency}, a strong robustness tuning method, to train a source model on XLM-RoBERTa-base. All backbone models are pretrained on the SQuAD\cite{rajpurkar-etal-2016-squad}, which serves as the source domain, using the default configuration provided by XTREME \cite{hu2020xtreme}.
In the sentiment analysis task, we pre-trained the Bert-base~\cite{devlin-etal-2019-bert} backbone using the SST-2~\cite{socher-etal-2013-recursive} dataset. Under the CTTA setting, we sequentially adapt each method to the sentiment analysis data stream.
During the CTTA phase, target test samples are sequentially fed into the model following the designed input order. At each time step $t$, the model is exposed solely to the current target test data. Each input sample is first evaluated and then immediately used for adaptation, with the student model’s prediction adopted as the final prediction. 
Our method is implemented using PyTorch, and all experiments are conducted on an NVIDIA N800 GPU. We set the batch size to 16 and utilize the AdamW optimizer \cite{loshchilov2018decoupled}.
The learning rate is uniformly set to 1e-5 across all methods. For our approach, the hyperparameter $\gamma$ is set to 0.4; $\tau$ is set to 1.2 when using XLMR-Base and 0.6 when using XLMR-Base-xTune. Baseline hyperparameters follow the default hyperparameters reported in the original works. All experiments are repeated three times with different random seeds, and the average results are reported.
Importantly, as there is no existing work on CTTA in NLP, all baseline methods are adapted from closely related TTA approaches. Specifically, Tent and OIL—originally designed for test-time adaptation on a fixed target domain—are modified to operate on the long and short sequence target task streams (Order 1 to 6) defined above. 
For CoTTA, which was originally developed for CTTA in computer vision, we adapt the task from image classification to text understanding and replace the image-based augmentation with text-based augmentation, such as random token replacement and deletion.

\noindent\textbf{Metrics.}
\textit{EM} follows a strict criterion: a prediction is considered correct if it exactly matches the ground truth (i.e., a character-level exact match); otherwise, it is deemed incorrect. \textit{F1} score, being the harmonic mean of Precision and Recall, offers a balanced measure of the model’s performance. 
EM emphasizes exact correspondence between the predicted answer and the reference answer, where a prediction is considered correct only if it matches the ground truth at the character level. This reflects the model’s ability to achieve strict semantic alignment. In contrast, the F1 captures partial overlaps between the prediction and the reference. Even if the prediction is not entirely correct, it can still receive credit as long as it contains key information, making F1 more tolerant to minor discrepancies.


\section{Tasks and CTTA Benchmark}
\label{app:task}
We establish a CTTA benchmark for text understanding across multiple domains and adaptation orders, covering three tasks:
Robust QA \cite{ravichander-etal-2021-noiseqa}, Reading Comprehension \cite{fisch2019mrqa}, and Cross-lingual QA \cite{artetxe2020cross}. The Robust QA task has two subsets (natural and synthetic noise), each with three subtasks (speech recognition, keyboard, and translation errors). The Reading Comprehension task includes Search and Trivia subtasks, while Cross-lingual QA includes translations into Chinese, Arabic, and Spanish. Each subtask is treated independently, resulting in 11 tasks. We construct short-sequence (7 tasks) and long-sequence (11 tasks) target task streams to evaluate performance under CTTA.
To construct long and short sequence target task streams, we use the following method: For the long-sequence setting (Orders 4 to 6), we organize the three task categories—Robust QA, Reading Comprehension, and Cross-lingual QA—into a fixed order (Order 4), numbering tasks 1 to 11. We then apply the Fisher-Yates shuffle \cite{liu2000statistical} twice to generate two randomized long-sequence streams (Orders 5 and 6). The short-sequence stream (Order 1) and its two randomized variants (Orders 2 and 3) are constructed similarly, excluding Robust QA-na and the Trivia subtask from Reading Comprehension. Details of Orders 1 to 6 are provided in Tab.~\ref{tab:task_sequences}.

The CTTA benchmark of sentiment analysis task includes cross-lingual settings (English, Chinese, Arabic, Spanish, etc.) and cross-domain settings (movie reviews, restaurant reviews, hotel reviews, travel reviews, written reviews, etc.) \cite{maas-etal-2011-learning,muennighoff-etal-2023-mteb,enevoldsenmmteb}. 

\begin{table}[htbp]
    \centering
    \caption{\textcolor{black}{The results of our method and baselines on the sentiment analysis data stream.}}
    \label{tab:sentiment_analysis}%
    \resizebox{1\linewidth}{!}{
    \begin{tabular}{lcccccccc}
    \toprule    
    \multicolumn{1}{l}{Time $\rightarrow$} & \multicolumn{1}{c}{imdb} & \multicolumn{1}{c}{eng} & \multicolumn{1}{c}{zho} & \multicolumn{1}{c}{ara} & \multicolumn{1}{c}{spa} & \multicolumn{1}{c}{jpn} & \multicolumn{1}{c}{rus} & \multicolumn{1}{c}{Avg} \\
 
    Method & Acc & Acc & Acc & Acc & Acc & Acc & Acc & Acc \\
    \midrule
    base & 87.64 & 92.01 & 58.71 & 50.63 & 77.05 & 44.08 & 64.81 & 70.94  \\
    CoTTA & 87.63 & 90.26 & 54.31 & 48.96 & 72.31 & 34.62 & 58.02 & 67.17  \\
    SoTTA & 87.69 & 92.03 & 53.96 & 48.84 & 74.52 & 35.62 & 57.46 & 67.73  \\
    SAR & 86.75 & 91.59 & 54.58 & 49.45 & 76.95 & 39.86 & 61.57 & 69.12  \\
    RAcc & 87.02 & 91.25 & 52.98 & 47.63 & 72.13 & 34.91 & 55.12 & 66.61  \\
    CTTA-T (ours) & \textbf{89.95} & \textbf{92.12} & \textbf{59.45} & \textbf{51.06} & \textbf{78.38} & \textbf{59.26} & \textbf{65.02} & \textbf{73.42}  \\
  
    \bottomrule
    \end{tabular}
    }
\end{table}
%

\textcolor{black}{
\begin{table}[htbp]
\centering
\caption{\textcolor{black}{Comparison with LLMs that exhibit strong zero-shot performance}}
\label{tab:zero-shot}
\resizebox{1\linewidth}{!}{
\begin{tabular}{lcccc}
\toprule
\textbf{Method} & \textbf{CTTA-T (ours)} & \textbf{GPT-4 Turbo} & \textbf{o3} & \textbf{Gemini 2.5 Pro} \\
\midrule
Order 1-6 (EM) & \textbf{57.64} & 53.03 & 56.39 & 57.31 \\
\bottomrule
\end{tabular}
}
\end{table}
\section{Comparison with LLMs that Exhibit Strong Zero-shot Performance}\label{app:zero-shot}
We report zero-shot LLM results on benchmarks in Tab.~\ref{tab:zero-shot}. The results show that even strong LLMs struggle when directly applied in a zero-shot manner to the challenging benchmarks considered in CTTA. By contrast, our method explicitly leverages online adaptation and is able to maintain substantially higher accuracy throughout domain shift.} 

\textcolor{black}{
Although LLMs demonstrate strong zero-shot capabilities, this ability fundamentally relies on their static pre-trained knowledge. In reality, knowledge evolves rapidly: new domain knowledge, specific noise patterns, and language usage continually emerge. Once trained, LLM parameters are frozen, making it difficult to cope with such fast-changing domain knowledge.}   

\textcolor{black}{
In contrast, CTTA provides an online self-adaptation mechanism that continuously leverages unlabeled test-stream data, allowing models to adapt in real time to newly emerging domains. The CTTA overcomes the limitations of static pre-trained models, enabling good performance even in unseen or rapidly changing environments.  }
\textcolor{black}{
\section{Time Cost of Each Module}
\label{app:cost_ablation}
We conduct a time cost experiment of each module, and the results are shown in Tab.~\ref{tab:cost_ablation},
where the \textbf{w/o IPCA \& w/o RFP} mean ablations on the cost introduced by the RFP module (Sec.~\ref {sec:3.2}) and the CDA module (Sec.~\ref{sec:3.3}).
The \textbf{ No adaptation} means the backbone model without any adaptation. 
The results show that: 
1) IPCA and RFP are computationally lightweight, increasing latency only marginally (<17\%). Importantly, both IPCA and RFP contribute significantly to performance improvements (see Tab.~\ref{tab: main_table}). Thus, their small computational cost is well justified.
2) CTTA-T introduces only a small overhead compared with the non-adapted backbone (<35\%), which is acceptable.
Overall, CTTA-T maintains low cost while delivering strong performance, confirming its practicality for real-time and online test-time settings.}
\begin{table}[htbp]
    \centering
    \caption{\textcolor{black}{The cost results of the ablation experiment at the module level.}}
    \label{tab:cost_ablation}
    \resizebox{1\linewidth}{!}{
    \begin{tabular}{lcccc}
    \toprule
     & \multicolumn{1}{c}{CTTA-T} & \multicolumn{1}{c}{w/o IPCA} & \multicolumn{1}{c}{w/o RFP} & \multicolumn{1}{c}{No adaptation} \\
    Backbone & Time (s) & Time (s) & Time (s) & Time (s) \\
    \midrule
    Base & 0.043 & 0.037 & 0.036 & 0.032 \\
    xTune & 0.040 & 0.036 & 0.034 & 0.032 \\
    \bottomrule
    \end{tabular}
    }
\end{table}

\section{Further Analysis Study of Error Accumulation}
\label{app:futher Collapse}
\begin{figure*}[htbp]
    \centering
    \includegraphics[width=1\textwidth]{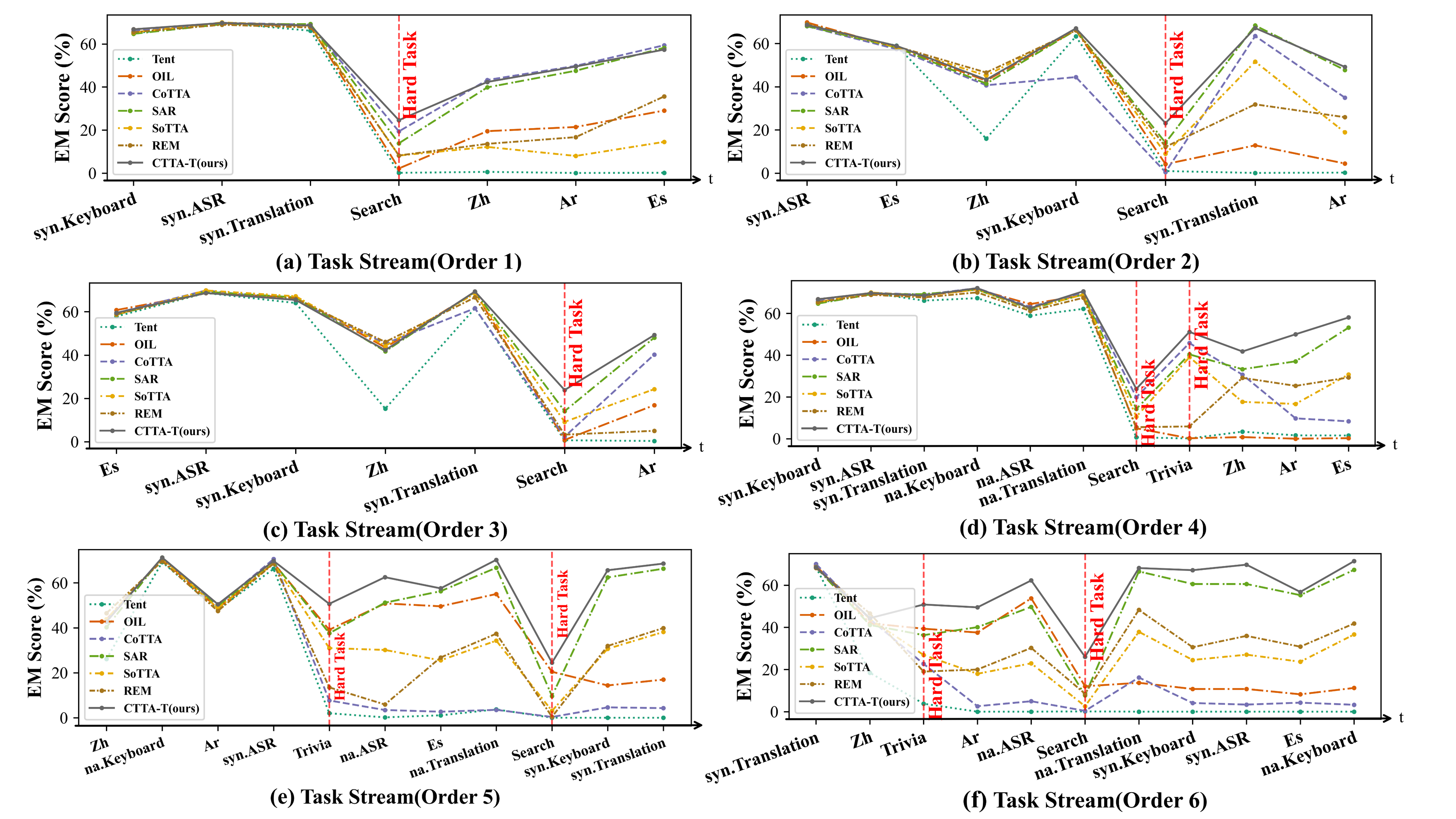}  
    \caption{
    Further Analysis Study of Model Collapse. The EM variations of each method under continuous domain shifts from Order 1 to Order 6 (excluding order 5)
    }
    \label{fig:big collapse}
\end{figure*}

To further analyze error accumulation, we compare our method with baselines on both short and long task sequences, with an emphasis on hard tasks (e.g., Search and Trivial) where accumulated errors are most likely to dominate. Specifically, the short sequence includes Search, while the long sequence includes both Search and Trivial. As shown in Fig.~\ref{fig:big collapse}, we report the EM variations of each method under continuous domain shifts from Order 1 to Order 6 (excluding Order 5).

All baselines exhibited severe error accumulation on hard tasks, whereas our method remained stable or showed only minimal degradation, highlighting its robustness in CTTA. 
OIL, SAR, SoTTA, and REM suffered from progressive error accumulation due to noisy and degrading pseudo-labels, which irreversibly propagated over time. 
Tent, by involving all samples in its entropy minimization, rapidly degenerated into a trivial solution where outputs were biased toward specific classes, leading to the earliest failure.
CoTTA deteriorated even earlier in Order 2, as its recovery strategy overwrote sensitive parameters, thereby accelerating error accumulation.

\section{Analysis Study of Backbone Influence}
\label{app:backbone_sensitivity}

Since the teacher model is initialized from a pretrained backbone, its quality may influence the overall CTTA performance. To study this factor, we evaluate different methods under two types of backbone initialization: a standard pretrained model (base) and a robustness-enhanced model (xTune). The results are summarized in Tab.~\ref{tab: main_table}. 
We observe that our method achieves consistently stable performance across both backbones. In contrast, other CTTA baselines perform unstably when using the weaker \texttt{base} backbone, and their results become relatively more stable with \texttt{xTune}, but still lag behind our method. This demonstrates that a stronger backbone indeed improves the quality of pseudo-labels and benefits all approaches, but our method is less sensitive to the initial teacher quality and remains reliable even with a weaker backbone.

\section{Analysis Study of the Stability of Each Method}
\label{variance}
To further assess stability under different adaptation orders, we report the variance of EM and F1 in Table~\ref{tab:Variance}. The variance is computed within each short-sequence (Orders 1–3) and long-sequence (Orders 4–6) setting, reflecting the sensitivity of each method to order variations. A smaller variance indicates more consistent performance across different orders. Our method (CTTA-T) consistently achieves the lowest variance across both metrics and sequence types, while baseline methods often display substantially larger variances. These results confirm that our method maintains stable adaptation behavior and is less affected by the sequence order, thereby validating its robustness in CTTA.
\begin{table*}[htbp]
\centering
\caption{Variance across long and short sequences. We report the variance of three adaptation orders within each sequence, calculated on a per-sequence basis. \textbf{Bold}: the smallest variance.}
\label{tab:Variance}
\resizebox{\textwidth}{!}{
\begin{tabular}{ccccccccccccccc}
\toprule
\multicolumn{1}{c}{} & \multicolumn{13}{c}{\textbf{Short Seq}} \\
\midrule
 & \multicolumn{2}{c}{\textbf{CTTA-T (ours)}} & \multicolumn{2}{c}{\textbf{REM}}& \multicolumn{2}{c}{\textbf{SoTTA}}& \multicolumn{2}{c}{\textbf{SAR}} & \multicolumn{2}{c}{\textbf{CoTTA}} & \multicolumn{2}{c}{\textbf{OIL}}  & \multicolumn{2}{c}{\textbf{Tent}} \\
\midrule
\textbf{Method} & \textbf{V(EM)} & \textbf{V(F1)} &  \textbf{V(EM)} & \textbf{V(F1)} & \textbf{V(EM)} & \textbf{V(F1)} & \textbf{V(EM)} & \textbf{V(F1)} & \textbf{V(EM)} & \textbf{V(F1)}& \textbf{V(EM)} & \textbf{V(F1)}& \textbf{V(EM)} & \textbf{V(F1)} \\
\midrule
\textbf{base} & \textbf{0.017} & \textbf{0.081} & 6.004 & 6.954 & 0.087 & 1.894 & 33.396 & 38.172 & 16.804 & 13.398 & 16.804 & 37.931 & 18.497 & 48.961 \\

\textbf{xtune} & \textbf{0.005} & \textbf{0.001} & 23.050 & 18.332 & 0.100 & 0.106 & 38.785 & 37.577 & 0.537 & 0.860 & 16.050 & 12.628  & 8.186 & 10.392 \\
\midrule
\multicolumn{1}{c}{} & \multicolumn{13}{c}{\textbf{Long Seq}} \\
\midrule
 & \multicolumn{2}{c}{\textbf{CTTA-T (ours)}} & \multicolumn{2}{c}{\textbf{REM}} & \multicolumn{2}{c}{\textbf{SoTTA}}& \multicolumn{2}{c}{\textbf{SAR}} & \multicolumn{2}{c}{\textbf{CoTTA}} & \multicolumn{2}{c}{\textbf{OIL}}  & \multicolumn{2}{c}{\textbf{Tent}} \\
\midrule
\textbf{Method} & \textbf{V(EM)} & \textbf{V(F1)} & \textbf{V(EM)} & \textbf{V(F1)} & \textbf{V(EM)} & \textbf{V(F1)} & \textbf{V(EM)} & \textbf{V(F1)} & \textbf{V(EM)} & \textbf{V(F1)}& \textbf{V(EM)} & \textbf{V(F1)}& \textbf{V(EM)} & \textbf{V(F1)} \\
\midrule
\textbf{base} & \textbf{0.003 } & \textbf{0.003} & 0.761 & 0.530 & 1.478 & 1.336 & 48.946 & 45.872 & 181.460 & 224.987 & 40.462 & 56.005  & 129.879 & 164.758 \\
\textbf{xtune} & \textbf{0.002} & \textbf{0.012 } & 7.361 & 4.324 & 1.157 & 0.862 & 15.319 & 16.303 & 0.365 & 0.312 & 26.798 & 40.776 & 179.894 & 245.460  \\
\bottomrule
\end{tabular}
}
\end{table*}

\section{Ablation Study of Effect of Each Component}
\label{app:Effect of Each Component}
In Tab.~\ref{tab:ablation_standard}, we assess the contribution of each component by removing it from the full model. Replacing the CDA module (Sec.~\ref{sec:3.4}) with a fixed EMA weight ($\alpha$ = 0.99) leads to a performance drop, indicating that dynamically adjusting EMA weights in response to domain shift enhances adaptation in CTTA settings. Similarly, removing the RFP module results in degraded performance, suggesting that the RFP module improves pseudo-label quality and effectively filters noisy samples. Notably, removing the SRT module causes the most substantial decline in performance, demonstrating its crucial role in breaking the chain of noise-induced erroneous updates and improving generalization.

\begin{table*}[htbp]
  \centering
  \caption{Ablation study on the short and long sequence benchmarks.
w/o CDA, w/o RFP, and w/o SRT indicate the removal of the corresponding modules from our framework. \textbf{Bold}: the best results.}
  \label{tab:ablation_standard}
    \resizebox{\textwidth}{!}{
  \begin{tabular}{lcccccccccccccc}
    \toprule
        \toprule
        \multicolumn{1}{l}{} & \multicolumn{6}{c}{\textbf{Short Sequence}} & \multicolumn{6}{c}{\textbf{Long Sequence}} &  \\
        \midrule
      \textbf{Order}& \multicolumn{2}{c}{\textbf{Order1 (\textbf{t}$\longrightarrow$)}} & \multicolumn{2}{c}{\textbf{Order2 (\textbf{t}$\longrightarrow$)}} & \multicolumn{2}{c}{\textbf{Order3 (\textbf{t}$\longrightarrow$)}} & \multicolumn{2}{c}{\textbf{Order4} (\textbf{t}$\longrightarrow$)} & \multicolumn{2}{c}{\textbf{Order5} (\textbf{t}$\longrightarrow$)} & \multicolumn{2}{c}{\textbf{Order6} (\textbf{t}$\longrightarrow$)} &\multicolumn{2}{c}{\textbf{Avg}}  \\
        \midrule
    base & EM & F1 & EM & F1 & EM & F1 & EM & F1 & EM & F1 & EM & F1 & EM & F1 \\
    \midrule
    \textbf{CTTA-T (ours) }     & \textbf{54.14} & \textbf{66.57} & \textbf{53.88} & \textbf{66.01} & \textbf{54.07} & \textbf{66.12} & \textbf{57.74} & \textbf{69.33} & \textbf{57.63} & \textbf{69.33} & \textbf{57.71} & \textbf{69.24} & \textbf{55.86} & \textbf{67.77} \\
    w/o CDA           & 53.51 & 65.83 & 53.40 & 65.55 & 53.14 & 65.40 & 57.30 & 68.86 & 57.43 & 68.95 & 57.36 & 68.83 & 55.36 & 67.24 \\
    w/o RFP           & 53.66 & 65.94 & 53.69 & 66.12 & 53.72 & 65.97 & 57.50 & 68.09 & 57.45 & 68.94 & 57.30 & 68.84 & 55.55 & 67.32 \\
    w/o SRT           & 48.12 & 59.35 & 46.37 & 57.29 & 51.22 & 62.78 & 48.71 & 58.74 & 51.90 & 63.17 & 39.46 & 49.76 & 47.63 & 58.52 \\
    \midrule
    xTune & EM & F1 & EM & F1 & EM & F1 & EM & F1 & EM & F1 & EM & F1 & EM & F1 \\
    \midrule
    \textbf{CTTA-T (ours) }     & \textbf{57.85} & \textbf{69.72} & \textbf{57.99} & \textbf{69.76} & \textbf{58.00} & \textbf{69.80} & \textbf{60.84} & \textbf{72.08} & \textbf{60.92} & \textbf{72.18} & \textbf{60.86} & \textbf{71.96} & \textbf{59.41} & \textbf{70.92} \\
    w/o CDA           & 57.53 & 69.60 & 57.69 & 69.59 & 57.92 & 69.59 & 60.33 & 71.50 & 60.88 & 71.92 & 60.58 & 71.72 & 59.16 & 70.65 \\
    w/o RFP           & 57.42 & 69.41 & 57.42 & 69.49 & 57.65 & 69.46 & 60.52 & 71.75 & 60.74 & 71.83 & 60.58 & 71.80 & 59.06 & 70.62 \\
    w/o SRT           & 55.12 & 66.58 & 55.45 & 66.85 & 57.22 & 68.85 & 55.36 & 65.71 & 57.46 & 68.49 & 53.68 & 64.58 & 55.72 & 66.84 \\
    \bottomrule
  \end{tabular}
  }
\end{table*}

\begin{figure*}[htbp]
    \centering
    \includegraphics[width=\textwidth]{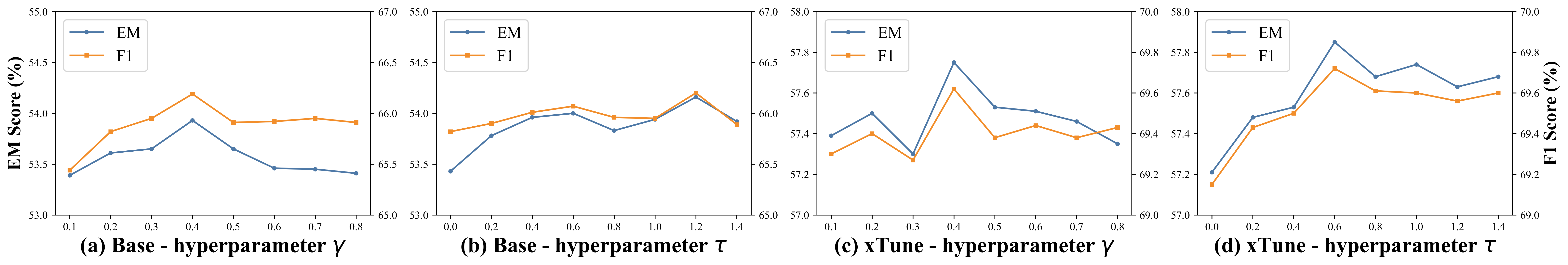}  
    \caption{
        Effect of different hyperparameters $\gamma$ and $\tau$ in the RFP module.
    }
    \label{fig:hyperpara}
\end{figure*}
\begin{table*}[htbp]
  \centering
  \caption{Performance of our method and a strong baseline, Anti-CF, on Orders 1 - 6. \textbf{Bold}: the best results.}
  \label{tab:anticf}
  \resizebox{\textwidth}{!}{%
    \begin{tabular}{lcccccccccccccc}
        \toprule
        \multicolumn{1}{l}{} & \multicolumn{6}{c}{\textbf{Short Sequence}} & \multicolumn{6}{c}{\textbf{Long Sequence}} &  \\
        \midrule
         \textbf{Order}& \multicolumn{2}{c}{\textbf{Order1 (\textbf{t}$\longrightarrow$)}} & \multicolumn{2}{c}{\textbf{Order2 (\textbf{t}$\longrightarrow$)}} & \multicolumn{2}{c}{\textbf{Order3 (\textbf{t}$\longrightarrow$)}} & \multicolumn{2}{c}{\textbf{Order4} (\textbf{t}$\longrightarrow$)} & \multicolumn{2}{c}{\textbf{Order5} (\textbf{t}$\longrightarrow$)} & \multicolumn{2}{c}{\textbf{Order6} (\textbf{t}$\longrightarrow$)} &\multicolumn{2}{c}{\textbf{Avg}}  \\
        \midrule
    Methods & EM & F1 & EM & F1 & EM &F1 & EM & F1 & EM & F1 & EM & F1 & EM & F1 \\
    \midrule
    base &       &       &       &       &       &       &       &       &       &       &       &       & 54.23 & 66.86 \\
    \hdashline\noalign{\vskip 1mm}
    Anti-CF & 52.83 & 65.57 & 53.35 & 65.69 & 53.23 & 65.64 & 56.47 & 68.50 & 56.22 & 68.37 & 56.18 & 68.32 & 54.71 & 67.02 \\
    \textbf{CTTA-T (ours)} & \textbf{54.14} & \textbf{66.57} & \textbf{53.88} & \textbf{66.01} & \textbf{54.07} & \textbf{66.12} & \textbf{57.74} & \textbf{69.33} & \textbf{57.63} & \textbf{69.33} & \textbf{57.71} & \textbf{69.24} & \textbf{55.86} & \textbf{67.77} \\
    \midrule
    xTune &       &       &       &       &       &       &       &       &       &       &       &       & 58.45 & 70.21 \\
    \hdashline\noalign{\vskip 1mm}
    Anti-CF & 56.47 & 68.94 & 56.82 & 68.96 & 56.67 & 68.95 & 59.73 & 71.49 & 59.47 & 71.19 & 59.54 & 71.11 & 58.12 & 70.11 \\
    \textbf{CTTA-T (ours)} & \textbf{57.85} & \textbf{69.72} & \textbf{57.99} & \textbf{69.76} & \textbf{58.00} & \textbf{69.80} & \textbf{60.84} & \textbf{72.08} & \textbf{60.92} & \textbf{72.18} & \textbf{60.86} & \textbf{71.96} & \textbf{59.41} & \textbf{70.92} \\
    \bottomrule
    \end{tabular}%
  }
\end{table*}

\begin{table*}[htbp]
  \centering
  \caption{The $p$-values of the t-test on our method.}
    \label{tab:t-test}
    \resizebox{\textwidth}{!}{
    \begin{tabular}{lcccccccccccc}
    \toprule
    \textbf{Order} & \multicolumn{2}{c}{\textbf{Order1}} & \multicolumn{2}{c}{\textbf{Order2}} & \multicolumn{2}{c}{\textbf{Order3}} & \multicolumn{2}{c}{\textbf{Order4}} & \multicolumn{2}{c}{\textbf{Order5}} & \multicolumn{2}{c}{\textbf{Order6}} \\
    \midrule
     base & EM & F1 & EM & F1 & EM & F1 & EM & F1 & EM & F1 & EM & F1 \\
    \midrule
    \textbf{CTTA-T} & 0.0001  & 0.0012  & 0.0002  & 0.0005  & 1.80e-5 & 0.0020  & 9.48e-5 & 0.0003  & 6.68e-8 & 0.0013  & 0.0003  & 0.0008  \\
    \midrule
     xtune & EM & F1 & EM & F1 & EM & F1 & EM & F1 & EM & F1 & EM & F1 \\
    \midrule
    \textbf{CTTA-T} & 7.08e-5 & 4.41e-6 & 2.34e-7 & 4.16e-5 & 2.78e-6 & 2.09e-5 & 0.0001  & 9.89e-5 & 6.74e-7 & 1.16e-5 & 1.27e-6 & 0.0006  \\
    \bottomrule
    \end{tabular}
    }
\end{table*}
\section{Strong Baseline Comparison}
\label{app:Strong_Baseline}
To further evaluate the adaptability of our method under the CTTA setting, we design a rigorous comparative experiment by selecting Anti-CF~\cite{su-etal-2023-beware}, a state-of-the-art method for test-time adaptation (TTA) in NLP, as a strong baseline. Anti-CF regularizes the adaptation process using outputs from the source model, which effectively mitigates the issue of model collapse commonly observed in TTA. However, it relies on a critical assumption: the availability of source training data during testing to warm up a side network—this fundamentally violates the CTTA principle of source-free test-time adaptation. 
Despite this violation, we include Anti-CF in our comparison to provide a comprehensive assessment of our method. As shown in Tab.~\ref{tab:anticf}, our method surpasses Anti-CF with an average improvement of 1.22\% in EM and 0.78\% in F1. Notably, when using xTune as the backbone model, Anti-CF even underperforms vanilla inference. This observation highlights the intrinsic difficulty of achieving stable adaptation under continually shifting domains, further demonstrating the robustness and practical value of our CTTA method.

\section{Hyperparameter Analysis}
\label{app:Hyperpara}
We conduct a hyperparameter search for $\gamma_c$ and $\tau$ in the RFP module (Sec. \ref{sec:3.3}) using Order. The threshold used in the main text is $\gamma_c$, and $\gamma_c=\gamma\log C$. Therefore, we only analyze $\gamma$. Since $\gamma$ and $\tau$ are interdependent, jointly searching them incurs a computational complexity of $\mathcal{O}(n^2)$. To address this, we first simplify the second-stage filtering in the RFP module. Specifically, instead of applying the \texttt{max} operation (i.e., $\max_j \pi_j = \pi_{\text{max}} < \gamma$), we sort $\pi(y^T)$ in descending order and compute the average of the top-$k$ values $\pi_{\text{avg}}$ that account for 30\% of the distribution mass. Samples with $\pi_{\text{avg}} < 0.2$ are filtered out. This modification approximates the model's consistency scoring across all samples as being uniform, effectively decoupling the influence of $\tau$ and enabling the isolated searching of $\gamma$.
After selecting an appropriate $\gamma$, we restore the original second-stage filtering of the RFP module and proceed with $\tau$ tuning. Fig.~\ref{fig:hyperpara} (a) and (b) present the EM and F1 scores results for different values of $\gamma$ and $\tau$ using the base backbone, and the optimal configuration is $\gamma = 0.4$ and $\tau = 1.2$. Similarly, Fig.~\ref{fig:hyperpara} (c) and (d) show the results for the xTune backbone, with the selected hyperparameters being $\gamma = 0.4$ and $\tau = 0.6$.

\section{Significance Test}
\label{app:t-test}
To assess whether the performance improvements achieved by our method are statistically significant across various evaluation metrics, we perform $t$-tests \cite{bartlett1937properties} on the main experimental results.
This test evaluates whether the differences in means between our method and baseline approaches are significant across different task orders. As shown in Tab.~\ref{tab:t-test}, the computed $p$-values for all orders are below 0.01, indicating that the observed performances are consistent and statistically significant rather than due to random chance.


\section{Societal Impacts}
\label{app:Social Impact}
The proposed CTTA-T framework addresses a critical challenge in real-world natural language processing systems: continual domain shift. By enabling the model to adapt continuously to changing test distributions, our method enhances the robustness and long-term viability of text understanding systems across various scenarios. These advancements could benefit users in multilingual, underrepresented, or rapidly changing environments, reducing performance degradation and decreasing dependence on expensive retraining or annotated data. However, adaptive models that update without explicit human control may become challenging to interpret and audit. While our method provides a solid foundation for continuous test-time adaptation in test understanding tasks, it is crucial to consider responsible and ethical usage policies in broader applications, particularly in areas involving sensitive data or high-risk decisions.

\end{document}